\documentclass[manuscript, screen, anonymous=false]{acmart}

\setcopyright{acmcopyright}
\copyrightyear{2023}
\acmYear{2023}
\acmDOI{xx.yyyy/zzzzzz.wwwwww}

\usepackage[utf8]{inputenc}
\usepackage{epsfig}
\usepackage{multirow}
\usepackage{booktabs}
\usepackage{dsfont}
\usepackage[compact]{titlesec}
\usepackage{tabularx}
\usepackage{xcolor,colortbl}
\usepackage{pdfpages} 
\usepackage{wrapfig}
\usepackage{subcaption}
\usepackage[textwidth=1.5cm]{todonotes}
\definecolor{kentuckyblue}{RGB}{0, 93, 170}

\excludecomment{oldcontent}

\usepackage{graphicx}
\newcommand{\STAB}[1]{\begin{tabular}{@{}c@{}}#1\end{tabular}}

\usepackage{natbib}
\newcommand{\reals}{\ensuremath{\mathbb{R}}}

\newcommand{\users}{\mathcal{U}}
\newcommand{\user}{u}
\newcommand{\items}{\mathcal{V}}
\newcommand{\itm}{v}
\newcommand{\features}{\phi}
\newcommand{\uprofile}{\omega}
\newcommand{\fagents}{\mathcal{F}}

\newcommand{\alloc}{\mathcal{A}}
\newcommand{\allochist}{\vec{H}}

\newcommand{\recfn}{\mathcal{R}}
\newcommand{\choice}{\mathcal{C}}
\newcommand{\choicehist}{\vec{L}}

\newcommand{\allocation}{\beta}
\newcommand{\prating}{\hat{r}}

\begin{document}

\title{Dynamic Fairness-aware Recommendation Through Multi-agent Social Choice}

\author{Amanda Aird}
\email{amanda.aird@colorado.edu}
\affiliation{%
  \institution{Department of Information Science, University of Colorado, Boulder}
  \city{Boulder}
  \state{Colorado}
  \country{USA}
  \postcode{80309}
}

\author{Paresha Farastu}
\email{paresha.farastu@colorado.edu}
\affiliation{%
  \institution{Department of Information Science, University of Colorado, Boulder}
  \city{Boulder}
  \state{Colorado}
  \country{USA}
  \postcode{80309}
}

\author{Joshua Sun}
\email{joshua.sun@colorado.edu}
\affiliation{%
  \institution{Department of Computer Science, University of Colorado, Boulder}
  \city{Boulder}
  \state{Colorado}
  \country{USA}
  \postcode{80309}
}

\author{Elena \u{S}tefancov\'{a}}
\email{elena.stefancova@fmph.uniba.sk}
\affiliation{%
  \institution{Comenius University Bratislava}
  \streetaddress{}
  \city{Bratislava}
  \country{Slovakia}
}

\author{Cassidy All}
\email{cassidy.all@colorado.edu}
\affiliation{%
  \institution{Department of Information Science, University of Colorado, Boulder}
  \city{Boulder}
  \state{Colorado}
  \country{USA}
  \postcode{80309}
}

\author{Amy Voida}
\email{amy.voida@colorado.edu}
\affiliation{%
  \institution{Department of Information Science, University of Colorado, Boulder}
  \city{Boulder}
  \state{Colorado}
  \country{USA}
  \postcode{80309}
}

\author{Nicholas Mattei}
\email{nsmattei@tulane.edu}
\affiliation{%
  \institution{Department of Computer Science, Tulane University}
  \city{New Orleans}
  \state{Louisiana}
  \country{USA}
  \postcode{70118}
}

\author{Robin Burke}
\email{robin.burke@colorado.edu}
\orcid{0000-0001-5766-6434}
\affiliation{%
  \institution{Department of Information Science, University of Colorado, Boulder}
  \city{Boulder}
  \state{Colorado}
  \country{USA}
  \postcode{80309}
}

\renewcommand{\shortauthors}{Aird, et al.}

\begin{abstract}
Algorithmic fairness in the context of personalized recommendation presents significantly different challenges to those commonly encountered in classification tasks. Researchers studying classification have generally considered fairness to be a matter of achieving equality of outcomes between a protected and unprotected group, and built algorithmic interventions on this basis. We argue that fairness in real-world application settings in general, and especially in the context of personalized recommendation, is much more complex and multi-faceted, requiring a more general approach. We propose a model to formalize multistakeholder fairness in recommender systems as a two-stage social choice problem. In particular, we express recommendation fairness as a novel combination of an allocation and an aggregation problem, which integrate both fairness concerns and personalized recommendation provisions, and derive new recommendation techniques based on this formulation. We demonstrate the ability of our framework to dynamically incorporate multiple fairness concerns using both real-world and synthetic datasets.
\end{abstract}


\maketitle

\section{Introduction}

Recommender systems are personalized machine learning systems that support users' access to information in applications as disparate as rental housing, video streaming, job seeking, social media feeds and online dating. The challenges of ensuring fair outcomes in such systems, including the unique issues that come from recommendation ecosystems, have been addressed in a growing body of research literature surveyed in several works including \citet{ekstrand2022fairness} and \citet{patro2022fair}. Despite these research efforts, some key limitations have remained unaddressed, including the dynamic and multi-stakeholder nature of recommendations systems, and these limitations render leave many solutions inadequate for the full range of applications for which recommender systems are deployed.

The first limitation we see in current work is that researchers have generally assumed that the problem of group fairness can be reduced to the problem of ensuring equality of outcomes between a protected and unprotected group, or in the case of individual fairness, that there is a single type of fairness to be addressed for all individuals. Fairness is a complex, multifaceted concept that can and does have different definitions, at different times, to different stakeholders, and these issues must be considered in the context in which the systems are deployed \cite{Hutchinson_2019,smith2023scoping}.  

We believe that this limitation is severe and not representative of realistic recommendation tasks in which fairness is sought. US anti-discrimination law, for example, identifies multiple protected categories relevant to settings such as housing, education and employment including gender, religion, race, age, and others \cite{Barocas2016-wi}. But even in the absence of such external requirements, it seems likely that any setting in which fairness is a consideration will need to incorporate the viewpoints of multiple groups.

We also expect that fairness will mean different things for different groups. Consider, for example, a system recommending news articles. Fairness might require that, over time, readers see articles that are geographically representative of their region: rural and urban or uptown vs downtown, for example. But fairness in presenting viewpoints might also require that any given day's set of headlines represent a range of perspectives. These are two different views of what fairness means, entailing different measurements and potentially different types of algorithmic interventions. The diversity of fairness definitions in a single system is rarely addressed: where fairness for multiple groups has been considered (e.g., \citet{sonboli2020opportunistic,kearns2018preventing}), it is defined in the same way for all groups.

The second limitation that we see in current work is that fairness-aware interventions in recommender systems, as well as many other machine learning contexts, have a static quality. In many applications, a system is optimized for some criterion and when the optimization is complete, it produces decisions or recommendations based on that learned state \cite{o2016weapons}. We believe it is more realistic to think of fairness as a dynamic state, especially when what is of primary concern are fair \emph{outcomes}. A recommender system's ability to produce outcomes that meet some fairness objective may be greatly influenced by context: what items are in inventory, what types of users arrive, how fair the most recent set of recommendations has been, among many others. A static policy runs the risk of failing to capitalize on opportunities to pursue fairness when they arise and/or trying to impose fairness when its cost is high, by not being sensitive to the context \cite{Burke:AlgoFairness}.

Our contribution in this paper is the design of an architecture for implementing fairness in recommender systems that addresses both of these limitations.\footnote{Note that portions of this work appeared in workshop form in \citet{burke2022multi}.} We start from the assumption that multiple fairness concerns should be active at any one time, and that these fairness concerns can be relatively unrestricted in form. Secondly, we build the framework to be \emph{dynamic}, in that decisions are always made in the context of historical choices and results.

Our research in fairness examines concepts inspired by the application context of Kiva Microloans, a non-profit organization which offers a platform (Kiva.org) for crowd-sourcing the funding of microloans, mostly in the developing world. Kiva's users (lenders) choose among the loan opportunities offered on the platform; microloans from multiple lenders that are aggregated and distributed through third party non-governmental organizations around the world. Kiva Microloans' mission specifically includes considerations of ``global financial inclusion''; as such, incorporating fairness in its recommendation of loans to potential users (lenders) is a key goal. The authors have been working with Kiva on a multi-faceted project addressing this goal and the work described here is, in part, an outgrowth of that project \cite{smith2023many,burke2022performance}. We will use Kiva's platform as an example throughout this paper. However, our findings and associated implementations are not specific to this setting. 

\section{Related Work}
There have been a number of efforts that explicitly consider the multisided nature of fairness in recommendation and matching platforms. \citet{patro2020fairrec} investigate fairness in two-sided matching platforms where there are both producers and consumers. They note, as we do, that optimizing fairness for only one side of the market can lead to very \emph{unfair} outcomes for the other side of the market. \citet{patro2020fairrec} also appeal to the literature on the fair allocation of indivisible goods from the social choice literature \cite{Thomson:FairRules}. They devise an algorithm that guarantees max-min share fairness of exposure to the producer side of the market and envy-free up to one item to the consumer side of the market \cite{amanatidis2023fair}. Their work is closest to the allocation phase of the system detailed in the following sections. However, in contrast to our work, they only use exposure on the producer side and relevance on the consumer side as fairness metrics, whereas our work aims to capture additional definitions. Also, we note that envy-freeness is only applicable when valuations are shared: a condition not guaranteed in a personalized system. It is possible for a user with unique tastes to receive low utility recommendations and still not prefer another user's recommendation lists. Also, our fairness formulation extends beyond the users receiving recommendations to providers of recommended items and envy-freeness provides no way to compare users who are getting different types of benefits from a system. In addition, our fairness definitions are dynamic, a case not considered by \citet{patro2020fairrec}. Indeed, in a recent survey paper, \citet{patro2022fair} provide an overview of some of the current work and challenges in fair recommendation, highlighting the focus of our work, multi-sided, multi-stakeholder, dynamic, and contextual issues, as key challenges.

Like \citet{patro2020fairrec}, the work of \citet{suhr2019two} investigates fairness in two-sided platforms, specifically those like Uber or Lyft where income opportunities are allocated to drivers. However, unlike our work and the work of \citet{patro2020fairrec}, \citet{suhr2019two} take proportionality as their definition of fairness, specifically proportionality with respect to time in a dynamic setting, and ensure that there is a fair distribution of income to the provider side of the platform. \citet{buet2022towards} also tackle the problem of multi-sided fair recommendation, arguing that recommender systems are built on sparse data and that the algorithms must take into account fairness. To this end they propose a regularization term that can be incorporated into the recommendation algorithm itself and account for various biases, e.g., exposure and/or selection bias, depending on the particular definition of the regularization term. The work of \citet{buet2022towards} is different from ours in that we make use of post-processing, as it allows for more flexibility w.r.t. the fairness definitions and allows us to treat the recommendation phase as a black-box, thus our system can be layered with any traditional recommendation algorithm. A post-processing approach also has the advantage that the set of fairness concerns and their relative importance can be adjusted on the fly without model retraining. 

Working with large item sets and multiple definitions of fairness can be computationally challenging. For example, \citet{zehlike2022fair} propose a system that allows for multiple protected groups on the provider side and their system works online, i.e., not in an offline, batch mode. But it is designed more for a general ranking task and is very computationally inefficient for recommendation. Every new recommendation list to be reranked requires the generation of a binary tree of size $|G|^k$, where $G$ is the number of protected groups and $k$ is the size of the output list. 

\citet{freeman2017fair} investigate what they call \emph{dynamic social choice functions} in settings where a fixed set of agents select a single item to share over a series of time steps. The work focuses on overall utility to the agents instead of considering the multiple sides of the recommendation interaction. Their problem is fundamentally a voting problem since all agents share the result, whereas we are focused on personalized recommendation. Their goal is to optimize the Nash Social Welfare of the set of agents (that remains fixed at each time step) and present four algorithms to find approximately optimal solutions. This work has a similar flavor to classical online learning / weighting experts problems \cite{cesa1997use} in the sense that the agent preferences remain fixed and the goal is to learn to satisfy them over a series of time steps. Similarly, \citet{kaya2020ensuring} focus on the problem of group recommendation, which is similar to classical social choice setting of multi-winner voting \cite{Zwicker:Voting}; i.e., a group of $k$ items are recommended as a shared resource to a set of users, and fairness is defined w.r.t. to this group of users \emph{preferences}. \citet{kaya2020ensuring} propose methods that are similar to those found in the multi-winner voting literature that attempt to be fair w.r.t. the individual preferences of the group, and these algorithms are meant to aggregate user preferences, not deliver specific recommendations on a per user basis, a key difference with our work.

\citet{ge2021towards} investigate the problem of long term dynamic fairness in recommendation systems. This work, like ours, highlight the need to ensure that fairness is ensured as a temporal concept and not as a static, one off, decision. To this end they propose a framework to ensure fairness of exposure to the producers of items by casting the problem as a constrained Markov Decision Process where the actions are recommendations and the reward function takes into account both utility and exposure. \citet{ge2021towards} propose a novel actor-critic deep reinforcement learning framework to accomplish this task at the scale of large recommender systems with very large user and item sets. Again, this work fixes definitions of fairness a priori, although their learning methodology may serve as inspiration to our allocation stage problems in the future. The subject of the long term impacts of recommender systems is also considered by \citet{akpinar2022long} who investigate the effects of various fairness interventions on the whole system, over time, finding similar effects to those of \citet{ge2021towards}. This reinforces the decisions in our model which take into account the long term, dynamics effects of any fairness intervention.

Reinforcement learning and other online learning methodologies have been proposed for various settings of fair recommendation. \citet{zhang2021recommendation} provide a short position paper proposig using reinforcement learning and an underlying Markov Decision process in order to learn priorities and feedback online during user interaction. While this is similar to a learning to rank system there are some key differences in their proposal, including learning fairness metrics. While Zhang and Wang do not propose a system, we agree that it is an intriguing direction and methodology to incorporate reinforcement learning into fair recommendation frameworks. In the learning to rank space, wherein a recommendation system is attempting to learn the users preference function online, like in RL, \citet{morik2020controlling} propose a new algorithm that is able to take into account notions of amortized group fairness while still learning user preferences. This is a fundamentally different setting than what we consider in that we are not performing learning to rank and we do not want to fix a set of fairness criteria a priroi into our recommendation algorithm, rather we treat the recommendation algorithm itself as an input.

\citet{morik2020controlling} investigate the problem of learning to rank over large item sets while ensuring fairness of merit based guarantees to groups of item producers. Specifically, they adapt existing methods to ensure that the exposure is \emph{unbiased}, e.g., that it is not subject to rich-get-richer dynamics, and \emph{fairness} defined as exposure being proportional to merit. Both of these goals are built into the regularization of the learner. In essence the goal is to learn user preferences while ensuring the above two desiderata. In contrast, our work factors out the recommendation methodology and we encapsulate the desired fairness definitions as separate agents rather than embedded in the learning algorithm. 

Similar to the RL settings described above, other recommendation systems contexts include session base and streaming (sequential) recommendation. \citet{wu2023faster} investigate the setting of \emph{session based recommender systems} which are short, memory-less recommendation experiences with, e.g., non-logged in users on a website. Like our work, they propose using the recommendation lists delivered over time in order to give the overall system some memory as to how fair it has been in the past, proposing a new time-decaying notion of exposure fairness. They also employ a post-processing concept for the overall recommendation: as they focus on session based recommender systems, there is no notion of long term user engagement. This is an important direction for potential future work, e.g., non-logged-in users are prevalent in many domains. Within the sequential recommender setting, \citet{li2022fairsr} propose a system for a (set of) users that incorporates feedback in learning both the preferences of the agent as well as a regularization term in the online learning algorithms to control for a type of interaction fairness per user. They use a deep learning framework work and a knowledge graph over the items to embed the user feedback and interaction and then use this to ensure that fairness of interaction is happening across protected item groups. Their model assumes fairness on a per-user (though dynamic) basis but more importantly that a large knowledge graph of item information is available, where we make no such assumptions and can define fairness both per-user and across use sets.

The architecture presented here advances and generalizes the approach found in \citet{sonboli2020and}. Like that architecture, fairness concerns are represented as agents and interact through social choice. However, in \citet{sonboli2020and}, the allocation mechanism selects only a single agent at each time step and the choice mechanism has a fixed, additive, form. We allow for a wider variety of allocation and choice mechanisms, and therefore present a more general solution. In addition to extending the work in \citet{sonboli2020and} by including the dynamic environment, we have worked with many different groups of stakeholders at Kiva.org in a qualitative research setting, allowing us to formalize a number of fairness concerns that operate at different levels of the recommendation ecosystem \cite{smith2023many}. Similarly, \citet{ferraro2021break} propose a novel set of re-rankers for a music recommendation platform after a set of interview studies with users and producers of music on the system. After these studies, they find that gender fairness is a key issue for many artists, with real world data they propose a re-ranking methodology that attempts to address this imbalance. The overall methods in this paper mirrors ours in that a close examination of a real world system, and interviews with stakeholders, give rise to a particular notion and measurement of fairness \cite{smith2023many}. However, rather than a case study, we propose a more robust framework that treats the multi-stakeholder and contextual definitions of fairness as first order concerns.

Finally, our recommendation allocation problem has some similarities with those found in computational advertising, where specific messages are matched with users in a personalized way~\citep{wang2017display,yuan2012internet}. Because advertising is a paid service, these problems are typically addressed through mechanisms of monetary exchange, such as auctions. There is no counterpart to budgets or bids in our context, which means that solutions in this space do not readily translate to supporting fair recommendation \citep{optimalbiding,edelman2007internet,yuan2013real}.

\section{Example}
In this section, we work through a detailed example demonstrating the function of the architecture through several iterations of user arrivals. The examples of fairness concerns articulated in our case study of Kiva.org arise from extensive interviews conducted with stakeholders at Kiva.org as a part of the larger scope of this project \cite{smith2023many}. Indeed, in our research we found that there were many competing definitions of fairness, e.g., proportional parity or minimal levels of exposure, that act on different sides of the market, i.e., the producers, consumers, or both. In what follows we give a detailed overview of building up the system from these concerns. The reader only interested in the formal definition of the overall system will find this in Section \ref{sec:formal_description}.

In short, we view the entire recommendation ecosystem as a \emph{multi-agent system}. Informally, each fairness concern can be represented by an agent, i.e., a collection of methods that is able to look at a (set of) recommendations and judge, for themselves, if these (potential) recommendations are \emph{fair} according to the definition of that agent. Likewise, users of the system can be viewed as a disjoint set of agents that only care if the provided recommendations conform to their interests, as judged by the user/item recommendation algorithm, i.e, the user preference. Given this view, we envision a system where, as a user agent arrives, one or more fairness agents are allocated to that user according to a user agent/fairness agent compatibility function. The agent is able to articulate a list that would be \emph{more fair}, i.e., provide a reranking methodology, and then in the aggregation stage we combine the lists from the fairness agent with the list generated by the user agent preferences alone. 

Given this setup, we have a multi-agent system where we are solving both a classical social choice allocation problem, i.e., allocating fairness agents to arriving users, and an aggregation (voting) problem, where we combine (possibly competing) lists of recommended items. In the following sections we articulate, in case study form, how one may approach the problem of formalizing multiple fairness concerns as \emph{agents} within our proposed system.

\subsection{Agents}
Consider the following set of fairness agents and their associated evaluations and preferences. We assume in this example that in all cases the agents' compatibility functions follow the pattern described in \citet{sonboli2020opportunistic} where the entropy of the user profile relative to the sensitive feature is calculated and users with high entropy are determined to be good targets for fairness-enhancing interventions. Note that in these examples, the definition of a fair outcome is differnet for each agent, demonstrating the range of fairness definitions that can be incorporated in our system:

\begin{description}
    \item [$f_{H}$: Health] This agent is concerned with promoting loans to the health sector. Its evaluation function compares the proportion of loans in the database in the health sector against the proportion of health recommendations in the recommendation list history. Its preference function is binary: if the loan is in the health sector, the score is 1; otherwise, zero. 
    
    \item [$f_A$: Africa] This agent is concerned with promoting loans to Africa. Its evaluation function, however, is list-wise. It counts lists in the recommendation if they have a least one loan recommendation to a country in Africa, and consider a fair outcome one in which every list has at least one such loan. Its preference function will be similarly binary as the $f_{H}$ agent.
    
    \item [$f_G$: Gender Parity] This agent is concerned with promoting gender parity within the recommendation history. If, across the previously generated recommendation lists, the number of men and women presented is proportional to their prevalence in the database, its evaluation will return 1. However, it is preference function is more complex than those above. If the women are underrepresented in the history, it will prefer loans to female borrowers, and conversely for men.\footnote{Note: At the time of our data gathering efforts, Kiva's borrower database recognized only binary gender categories.}
    
    \item [$f_L$: Large] This agent is concerned with promoting loans with larger total amounts: over \$5,000. Internal Kiva research has shown that such loans are often very productive because they go to cooperatives and have a larger local impact. However, the same research has shown that Kiva users are less likely to support them because each contribution has a smaller relative impact.\footnote{Pradeep Ragothaman, Personal communication} This agent is similar to the $f_{A}$ agent above in that it seeks to make sure each list has one larger loan. 

\end{description}

\subsection{Loans}

Consider the contents of Table~\ref{table:example-loans}. For the sake of example, we will assume these loans, characterized by the Region, Gender, Section and Amount, constitute the set of loans available for recommendation.

{\small
\begin{table}[tbh]
    \begin{tabular}{|c|c|c|c|c|}
    \hline
            & $\phi^s_{1}$ : Region & $\phi^s_{2}$ : Gender & $\phi^s_{3}$ : Sector & $\phi_{4}$ : Amount \\
    \hline
        $v_1$ & Africa & Male & Agriculture & \$5,000-\$10,000\\
    \hline
        $v_2$ & Africa & Female & Health & \$500-\$1,000\\
    \hline
        $v_3$ & Middle-East & Female & Clothing & \$0-\$500 \\
    \hline
        $v_4$ & Central America & Female & Clothing & \$5,000-\$10,000 \\
    \hline
        $v_5$ & Central America & Female & Health & \$0-\$500 \\
    \hline
        $v_6$ & Middle-East & Female & Clothing & \$0-\$500 \\
    \hline
    \end{tabular}
    \caption{Set of Potential Loans.}
    \label{table:example-loans}
\end{table}
}

\subsection{Mechanisms}
For the sake of exposition, we posit two very simple mechanisms for allocation and choice. We will assume that our allocation mechanism is a single outcome lottery, e.g., a randomized allocation mechanism \cite{budish2013designing}. One agent will be chosen to participate in the choice mechanism, based on a random draw with probabilities based on the historic unfairness and user compatibility as measured by each agent.

We assume that the recommendation lists are of size 3 and the choice mechanism uses a weighted voting / score-based mechanism \cite{BCELP16a} using a weighted sum of 0.75 on the personalized results for the recommender system and 0.25 on the output of the allocated fairness agent.

\subsection{Users}
At time $t_1$, \textbf{User $u_1$} arrives at the system and the recommendation process is triggered. The user has previously supported small loans only in Central America and Middle East, but has lent to a wide variety of sectors and genders.

For the sake of example, we will assume that 
the agents measure their prior history relative to their objectives as equally unfair at $0.5$, except the Gender Parity agent, which starts out at parity and therefore returns a value of 1. However, the compatibility functions for $f_A$ and $f_L$ returns lower scores because of the user's historical pattern of lending. This yields a lottery in which $f_G$ has probability zero, $f_A$ has a low probability, and $f_H$ a higher one. The allocation mechanism chooses randomly, and we will assume that $f_H$, the health-focused agent, is picked. 

The recommender returns the following  list of items and predicted ratings $[\{v_6, 0.6\}, \{v_4, 0.5\}, \{v_5, 0.3\}, \{v_3, 0.3\},$ $ \{v_1, 0.0\}, \{v_2, 0.0\}]$. The $f_H$ agent gives a score of 1 to the health-related loans $v_2$ and $v_5$ and 0 to all others. 

The choice mechanism combines these scores as described above and returns the final recommendation list $[\{v_5, 0.475\}, \{v_6, 0.45\}, \{v_4, 0.375\}]$. Note that the Health agent has successfully promoted its preferred item to the first position in the list. 

For the sake of example, we assume that the agents' evaluation functions are very sensitive. Therefore, when \textbf{User $u_2$} arrives, the results of the previous recommendations have caused the evaluations to shift such that the Health $f_H$ and Large $f_L$ agents are now satisfied (note that $v_4$ is included in $u_1$'s list and it was a large loan), the Gender parity agent $f_G$ is now at $0.9$ (note that there is only one male loan in the database) but the Africa agent $f_A$, which got nothing in $u_1$'s list, considers recent results to be unfair with a score of $0.25$. We assume that $u_2$ is similar to $u_1$ in profile and therefore compatibility, but $f_A$ has a much worse fairness score than $f_G$, and therefore a high allocation probability. We will assume $f_A$ is chosen.

Because this user has similar preferences to $u_1$, they get the same recommendations: $[\{v_6, 0.6\}, \{v_4, 0.5\}, \{v_5, 0.3\}, \{v_3, 0.3\}],$ $\{v_1, 0.0\}, \{v_2, 0.0\}]$. The $f_A$ agents scores the two loans from Africa ($v_1$ and $v_2$) at 1 and the others at 0. 

So, after randomly breaking the tie between $v_1$ and $v_2$, the final recommendation list is $[\{v_6, 0.45\}, \{v_4, 0.375\}, \{v_1, 0.25\}]$. 

When \textbf{User $u_3$} arrives, all four agents find themselves scoring fairness at 1 over the evaluation window and so no agents are allocated. The results from the recommendation algorithm pass through the choice mechanism unchanged and are delivered to the user.

In this example, we see the interplay between users' compatibility with agents and the computed fairness outcomes to allocate opportunities to pursue fairness among different agents over time.

\section{Formalizing Fairness Concerns}
A central tenet of our work is that fairness is a contested concept \cite{mulligan2019thing}. From an application point of view, this means that ideas about fairness will be grounded in specific contexts and specific stakeholders, and that these ideas will be multiple and possibly in tension with each other. From a technical point of view, this means that any fairness-aware recommender system should be capable of integrating multiple fairness concepts, arising as they may from this contested terrain.

A central concept in this work is the idea of a \textit{fairness concern}. We define a fairness concern as a specific type of fairness being sought, relative to a particular aspect of recommendation outcomes, evaluated in a particular way. As shown in the example above, a possible fairness concern in the microlending context might be group fairness relative to different geographical regions considered in light of the exposure of loans from these regions in recommendation lists.\footnote{We are currently conducting research to characterize fairness concerns appropriate to Kiva's recommendation applications. See \citet{smith2023many} for some initial findings from this work. None of the discussion here is intended to represent design decisions or commitments to particular concerns and/or their formulation.} The concern identifies a particular aspect of the recommendation outcomes (in this case, their geographical distribution), the particular fairness logic and approach (more about this below), and the metric by which fair or unfair outcomes are determined.  

The first consideration in building a fairness-aware recommender system is the question of what fairness concerns surround the use of the recommender system, itself. Many such concerns may arise and like any system-building enterprise, there are inevitably trade-offs involved in the formulation of fairness concerns. An organization may decide to incorporate only the highest-priority concerns into its systems. An initial step in fairness-aware recommendation is for an organization to consult its institutional mission and its internal and external stakeholders with the goal of eliciting and prioritizing fairness concerns. We report on our initial phases of stakeholder consultation with Kiva.org in \citet{smith2023many}. Although not in the recommendation domain, another relevant project is the WeBuildAI project~\cite{lee2019webuildai} and its participatory design framework for AI. 

In addition to addressing different aspects of system outcomes, different fairness concerns may invoke different logics of fairness. Welfare economists have identified a number of such logics and we follow \citet{moulin2004fair} who identifies four:
\begin{description}
    \item[Exogenous Right:] A fairness concern is motivated by exogeneous right if it follows from some external constraint on the system. For example, the need to comply with fair lending regulations may mean that male and female borrowers should be presented proportionately to their numbers in the overall loan inventory.
    \item[Compensation:] A fairness concern that is a form of compensation arises in response to observed harm or extra costs incurred by one group versus others. For example, as noted above, loans with longer repayment periods are often not favored by Kiva users because their money is tied up for longer periods. To compensate for this tendency, these loans may need to be recommended more often. 
    \item[Reward:] The logic of reward is operational when we consider that resources may be allocated as a reward for performance. For example, if we know that loans to large cooperative groups are highly effective in economic development, we may want to promote such loans as recommendations so that they are more likely to be funded and realize their promise. 
    \item[Fitness:] Fairness as fitness is based on the notion of efficiency. A resource should go to those best able to use it. In a recommendation context, it may mean matching items closely with user preferences. For example, when loans have different degrees of repayment risk, it may make sense to match the loan to the risk tolerance of the lender.
\end{description}

It is clear that fairness logics do not always pull in the same direction. The invocation of different logics are often at the root of political disagreements: for example, controversies over the criteria for college admissions sometimes pit ideas of reward for achievement against ideas of compensation for disadvantage.

Recommender systems often operate as two-sided platforms, where one set of individuals are receiving recommendations and possibly acting on those recommendations (consumers), and another set of individuals is creating or providing items that may be recommended (providers) \cite{burke_multisided_2017}. Consumers and providers are considered, along with the platform operator, to be the direct stakeholders in any discussion of recommender system objectives. Fairness concerns may derive from any stakeholder, and may need to be balanced against each other. The platform may be interested in enforcing fairness, even when other stakeholders are not. For example, the average recommendation consumer might only be interested in the best results for themselves, regardless of the impact on others. Fairness concerns can arise on behalf of other, indirect, stakeholders who are impacted by recommendations but not a party to them. An important example is \textit{representational fairness} where concerns arise about the way the outputs of a recommender system operate to represent the world and classes of individuals within it: for example, the way the selection of news articles might end up representing groups of people unfairly \cite{noble2018algorithms} (see \cite{ekstrand2022fairness} for additional discussion). As a practical matter, representational fairness concerns can be handled in the same way as provider-side fairness for our purposes here.

Finally, we have the consideration of group versus individual fairness. This dichotomy is well understood as a key difference across types of fairness concerns, defining both the target of measurement of fairness and the underlying principle being upheld. Group fairness requires that we seek fairness across the outcomes relative to predefined protected groups. Individual fairness asks whether each individual user has an appropriate outcome and assumes that users with similar profiles should be treated the same. Just as there are tensions between consumer and provider sides in fairness, there are fundamental incompatibilities between group and individual fairness. Treating all of the outcomes for a group in aggregate is inherently different than maintaining fair treatment across individuals considered separately. Friedler et al. offer a thorough discussion of this topic \cite{friedler2021possibility}. 

\begin{table*}[tbh]
{\small
\begin{tabular}{l|l|l|l|p{1.0in}|p{1.0in}}
    Label & Fairness type & Logic & Side & Who is Impacted & Evaluation \\ \hline
    LowCountry & Group & Comp. & Provider & Borrowers from countries with lower funding rates & Exposure of loans in recommendation lists \\ \hline
     LargeAmt & Group & Reward & Provider & Borrowers in consortia seeking larger loans & Exposure of loans in recommendation lists \\ \hline
     Repay & Individual & Reward & Provider & All borrowers & Loan exposure proportional to repayment probability \\ \hline
     LowSector & Group & Exo. right & Provider & Borrowers in sectors with lower funding rates & Exposure of loans in recommendation lists \\ \hline
     AllCountry & Individual & Exo. right & Provider & All borrowers & Catalog coverage by country \\ \hline 
     AccuracyLoss & Group & Exo. right & Consumer & All lenders & Accuracy loss due to fairness objective is fairly distributed across protected groups of users. \\ \hline
     RiskTolerance & Individual & Fitness & Consumer & All lenders & Riskier loans are recommended to users with greater risk tolerance
\end{tabular}
}
\caption{Potential fairness concerns and their logics.}
\label{tab:concerns}
\vspace{-0.5cm}
\end{table*}

Putting all of these dimensions together gives us a three-dimensional ontology of fairness concerns in recommendation: fairness logic, consumer- vs provider-side, and group vs individual target. Table \ref{tab:concerns} illustrates a range of different fairness concerns that derived from the microlending context and all of which have at least some support from the interview study by \citet{smith2023many}. This list illustrates a number of the points relative to fairness concerns raised so far. We can see that all four of Moulin's fairness logics are represented. We also see that the fairness concerns can be group or individual: for example, we are attentive to individual  qualities in the \textbf{RiskTolerance} concern, but group outcomes in  \textbf{LargeAmt}. 
The \textbf{AccuracyLoss} concern is a consumer-side concern, relevant to lenders, but other concerns are on the provider side. We also see that it is possible for a single objective, here the geographic diversity of loan recommendation, to be represented by multiple fairness concerns: \textbf{LowCountry} and \textbf{AllCountry}. In spite of having the same target, these concerns are distinguished because they approach the objective from different logics and evaluate outcomes differently. 

\subsection{Fairness Agents} \label{sec:fairness_agents}
Our architecture SCRUF-D (Social Choice for Recommendation Under Fairness -- Dynamic) \cite{burke2022multi} builds on the SCRUF architecture introduced in \citep{Burke:AlgoFairness,sonboli2020and}. It is designed to allow multiple fairness concerns to operate simultaneously in a recommendation context. Fairness concerns, derived from stakeholder consultation, are instantiated in the form of fairness agents, each having three capabilities:

\begin{description}
    \item [Evaluation:] A fairness agent can evaluate whether the current historical state is fair, relative to its particular concern. Without loss of generality, we assume that this capability is represented by a function $m_i$ for each agent $i$ that takes as input a history of the system's actions and returns an number in the range $[0,1]$ where 1 is maximally fair and 0 is totally unfair, relative to the particular concern.
    \item [Compatibility:] A fairness agent can evaluate whether a given recommendation context represents a good opportunity for its associated items to be promoted. We assume that each agent $i$ is equipped with a function $c_i$ that can evaluate a user profile $\uprofile$ and associated information and return a value in the range $[0,1]$ where 1 indicates the most compatible user and context and 0, the least.
    \item [Preference:] An agent can compute a preference for a given item whose presence on a recommendation list would contribute (or not) to its particular fairness concern. Again, without loss of generality, we assume this preference can be realized by a function that accepts an item as input and returns a preference score in $\reals_+$ where a larger value indicates that an item is more preferred.\footnote{A more complex preference scenario is one in which agents have preferences over entire lists rather than individual items. We plan to consider such preference functions in future work.}
\end{description}

\subsection{Recommendation Process}

We assume a recommendation generation process that happens over a number of time steps $t$ as individual users arrive and recommendations are generated on demand. Users arrive at the system one at a time, receive recommendations, act on them (or not), and then depart. When a user arrives, a recommendation process produces a recommendation list $\ell_s$ that represents the system's best representation of the items of interest to that user, generated through whatever recommendation mechanism is available. We do not make any assumptions about this process, except that it is focused on the user and represents their preferences. A wide variety of recommendation techniques are well studied in the literature, including matrix factorization, neural embeddings, graph-based techniques, and others. 

The first step to incorporating fairness into the recommendation process is to determine which fairness concerns / agents will be active in responding to a given recommendation opportunity. This is the \textit{allocation phase} of the process, the output of which is a set of non-negative weights $\beta$, summing to one, over the set of fairness agents, indicating to what extent each fairness agent is considered to be allocated to the current opportunity. 

Once the set of fairness agents have been allocated, they have the opportunity to participate in the next phase of the process, which is the \textit{choice phase}. In this phase, all of the active (non-zero weighted) agents and their weights participate in producing a final list of recommendations for the user. We view the recommender system itself as being an agent that is always allocated and therefore always participates in this phase. 

\section{The SCRUF-D Architecture}
The two phases of the SCRUF-D architecture are detailed in Figures \ref{fig:alloc} and \ref{fig:choice}. The original SCRUF framework \cite{sonboli2020and} concentrated on the representation of user preferences, as computed by the recommender system, and fairness concerns, as derived from stakeholder consultation as discussed in Section \ref{sec:fairness_agents}, and their integration. SCRUF-D incorporates the history of system decisions and the fairness achieved over time to control the allocation of fairness concerns. We will first provide a high level overview of the system and describe each aspect in detail with formal notation: Table~\ref{tab:notation} provides a reference to this notation.

\subsection{Overview}
We can think of a recommender system as a two-sided market in which the recommendation opportunities that arise from the arrival of a user $\user \in \users$ to the system, and each are allocated to a set of items $\itm \in \items$ from the system's catalog. This market has some similarities to various forms of online matching markets including food banks \citep{AAGW15a}, kidney allocation \citep{MaSaWa18Com,AwSa09a}, and ride sharing \citep{dickerson2018allocation}, in that users have preferences over the items; however, in our case this preference is known only indirectly through either the prior interaction history or a recommendation function. Additionally, the items are not consumable or rivalrous. For example, a loan can be recommended to any number of users -- it is not ``used up'' in the recommendation interaction.\footnote{Loans on Kiva's platform may be exhausted eventually through being funded, but many other objects of recommendation such as streaming media assets are effectively infinitely available.} Also, users are not bound to the recommendations provided; in most online platforms including Kiva, there are multiple ways to find items, of which the recommender system is only one. 

Once we have a collection of fairness agents we must solve two interrelated problems: 
\begin{enumerate}
    \item What agent(s) are allocated to a particular recommendation \emph{opportunity}?
    \item How do we \emph{balance} between the allocated agents and the user's individual preferences?
\end{enumerate}

\begin{figure*}[tb]
\vspace{-3mm}
\centering
\includegraphics[width=0.75\textwidth]{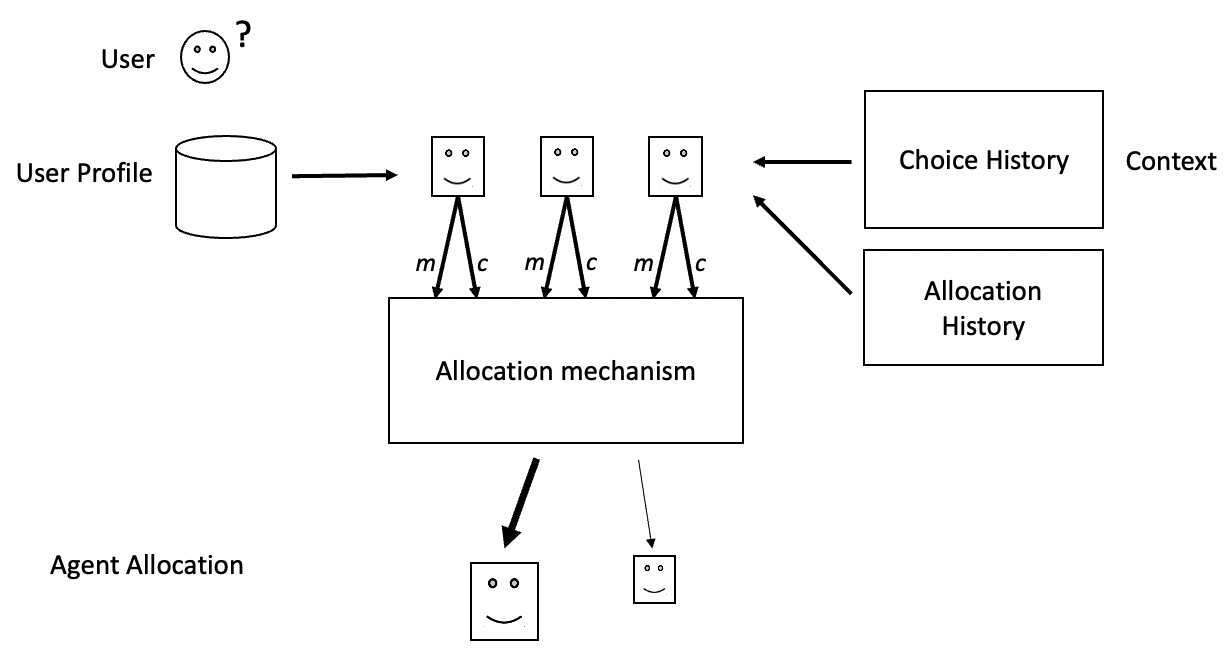}
\vspace{-3mm}
\caption{SCRUF-D Framework / Allocation Phase: Recommendation opportunities are allocated to fairness concerns based on the context.}
\label{fig:alloc}
\end{figure*}

Figure \ref{fig:alloc} shows the first phase of this process, allocation \cite{BCELP16a}, in which the system decides which fairness concerns / agents should be allocated to a particular fairness opportunity. This is an online and dynamic allocation problem where we may consider many factors including the history of agent allocations so far, the generated lists from past interactions with users, and how fair the set of agents believes this history to be. As described in Section~\ref{sec:fairness_agents}, agents take these histories and information about the current user profile and calculate two values: $m$, a measure of fairness relative to their agent-specific concern, and $c$, a measure of compatibility between the current context and the agent's fairness concern. The allocation mechanism takes these metrics into account producing a probability distribution over the fairness agents that we call the \emph{agent allocation}, which can be interpreted as weights in the choice stage or be used to select a single agent via a lottery, e.g., a randomized allocation scheme \cite{budish2013designing}.

\begin{figure*}[tb]
\vspace{-3mm}
\centering
\includegraphics[width=0.75\textwidth]{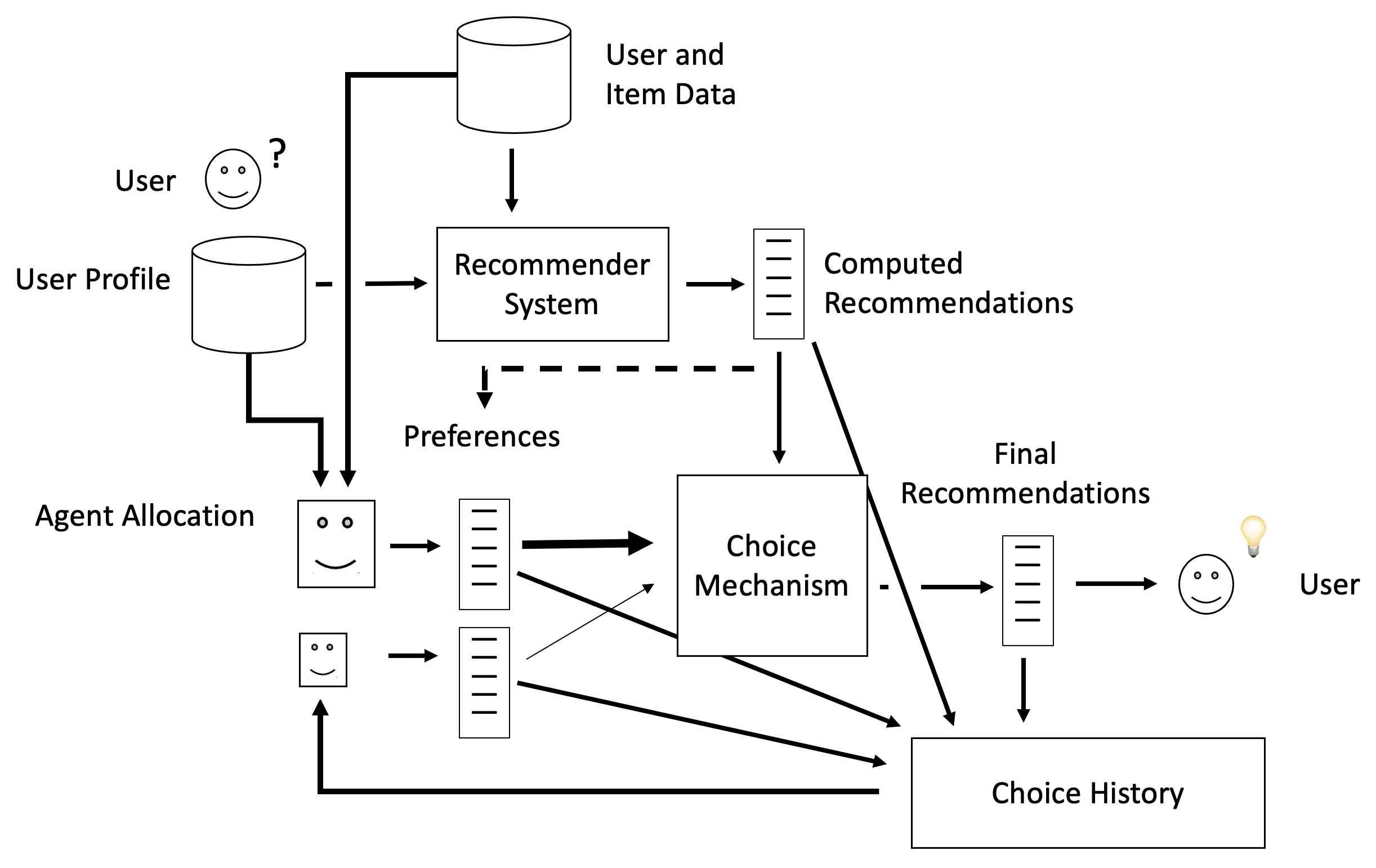}
\vspace{-3mm}
\caption{SCRUF-D Framework / Choice Phase: The preferences derived from the recommender system and the fairness concerns are integrated by the choice mechanism.}
\label{fig:choice}
\end{figure*}

In the second phase, shown in Figure \ref{fig:choice}, the recommender system generates a list of options, considered to represent the user's preferences. The fairness concerns generate their own preferences as well. These preferences may be global in character, i.e., preferences over all items, in which case they may be independent of what the recommender system produces; we call this a recommendation function below. Or, as indicated by the dashed line, these preferences may be scoped only over the items that the recommender system has generated; named a scoring function. In either case, the preference function of the fairness agent, like the one for the user, generates a list of items and scores.  The choice mechanism combines these preferences of both the user and fairness agents, along with the allocation weights of the fairness agents, to arrive at a final recommendation list to be delivered to the user. The list itself, and possibly the interactions the user has with it, becomes a new addition to the choice history and the process continues for the next user.

\begin{center}
\begin{table}[tbh]
{\small
    \begin{tabular}{|c|cp{7.2cm}|}
    \hline
            \multirow{8}{*}[-3ex]{\STAB{\rotatebox[origin=c]{90}{Rec. System}}} & $\users (\user)$ & Users (user).  \\
            & $\items (\itm)$ & Items (item). \\
            & $\features = \langle \features_{1}, \ldots \features_{k} \rangle$ & Item Features. \\
            & $\uprofile = \langle \uprofile_{1}, \ldots \uprofile_{j} \rangle$ & User Profile. \\
            & $\features^{s} \subseteq \features$ & Sensitive Item Features as a subset of all item features $\features$. \\
            & $\uprofile^{s} \subseteq \uprofile$ & Sensitive Aspects of User Profile as a subset of all user profile features $\uprofile$. \\
            & $\recfn_i(\uprofile, \itm) \rightarrow \{\itm, \prating \}$ & Recommendation mechanism that takes a user profile $\uprofile$ and a (set of) items $\itm$ and produces a predicted rating $\prating \in \reals_+$.  \\
            
            & $ \ell = \langle \{\itm_1, \prating_1\}, \ldots \{\itm_i, \prating_i\} \rangle $ & Recommendation List as an ordered list of item, predicted rating pairs. \\
            
            & $sort(\recfn_i(\uprofile, \items)) \rightarrow \ell$ & Recommendation List for user $\uprofile$ sorted by $\prating$. \\
    \hline
            \multirow{6}{*}[-7ex]{\STAB{\rotatebox[origin=c]{90}{Fairness Agents}}} & $\fagents = \{f_1, \ldots, f_i\}$ & Set of Fairness Agents. \\
            
            & $f_i = \{m_i, c_i, \recfn_i\}$ & Fairness agent $i$ defined by a fairness metric $m_i$, a compatibility metric $c_i$, and a ranking function $\recfn_i$. \\
            
            & $m_i(\choicehist, \allochist) \rightarrow [0,1]$ & Fairness metric for agent $i$ that takes a choice history $\choicehist$ and allocation history $\allochist$ and produces a value in $[0,1]$ according to the agent's evaluation of how fair recommendations so far have been.  \\
            
            & $c_i(\uprofile) \rightarrow [0,1]$ & Compatibility metric for agent $i$ that takes a particular user profile $\uprofile$ and produces a value in $[0,1]$ for how compatible fairness agent $i$ believes they are for user $\uprofile$. Note: The compatibility metric combines preferences on the agent side and those on the user side (inferred from the profile). If these preferences are symmetrical, we have a one-sided matching problem, but two-sided cases are also possible. \\
            
            & $\recfn_i(\uprofile, v) \rightarrow \{\itm, \prating\}$ & Fairness Agent Recommendation function.  \\
            
            & $\recfn_i(\ell, \uprofile, v) \rightarrow \{\itm, \prating\}$ & Fairness Agent Scoring function.  \\
            
            & $\ell_\fagents =  \{\recfn_1(\uprofile, \items), \ldots, \recfn_i(\uprofile, \items)\}$ & Set of Fairness Agent Recommendation Lists indexed by fairness agent label $i$.\\
    \hline
            \multirow{2}{*}[-4ex]{\STAB{\rotatebox[origin=c]{90}{Allocation}}} & $\alloc (\fagents, m_\fagents(\choicehist, \allochist), c_\fagents(\uprofile))
            \rightarrow \beta \in \reals_{+}^{|\fagents|} $  & Allocation mechanism $\alloc$ that takes a set of fairness agents $\fagents$, the agents' fairness metric evaluations $m_\fagents(\choicehist, \allochist)$, and the agents' compatibility metric evaluations $c_\fagents(\uprofile)$ and maps to an agent allocation $\allocation$.\\
            
            & $\allochist = \langle \allocation^1, \ldots, \allocation^t \rangle$ & Allocation History $\allochist$ that is an ordered list of agent allocations  $\alloc$ at time $t$. \\
    \hline
            
            \multirow{2}{*}[-4ex]{\STAB{\rotatebox[origin=c]{90}{Choice}}} & $\choice(\ell, \allocation, \ell_\fagents) \rightarrow \ell_\choice$ & Choice Function is a function from a recommendation list $\ell$, agent allocation $\allocation$, and fairness agent recommendation list(s) $\ell_\fagents$ to a combined output list $\ell_\choice$. \\
            
            & $\choicehist = \langle \ell^t, \ell^t_\fagents, \ell^t_C \rangle$ & Choice History that is an ordered list of user recommendation list $\ell$, agent recommendation list(s) $\ell_\fagents$, and choice function output lists $\ell_\choice$, indexed by time step $t$.   \\
            
        \hline
    \end{tabular}
}
    \caption{Notations for our formal description of the SCRUF-D architecture.}
    \label{tab:notation}
\end{table}
\end{center}

\subsection{Formal Description}\label{sec:formal_description}

In our formalization of a recommendation system setting we have a set of users $\users = \{\user_1, \ldots \user_n\}$ and a set of items $\items = \{\itm_1, \ldots, \itm_m\}$. For each item $\itm_i \in \items$ we have a $k$-dimensional feature vector $\features = \langle \features_{1}, \ldots \features_{k} \rangle$ over a set of categorical features $\features$, each with finite domain. Some of these features may be sensitive, e.g., they are associated with one or more fairness agent concerns, we denote this set as $\features^{s}$. Without loss of generality, we assume that all elements in $\items$ share the same set of features $\features$. Finally, we assume that each user is associated with a profile of attributes $\uprofile = \langle \uprofile_{1}, \ldots \uprofile_{j} \rangle$, of which some also may be sensitive $\uprofile^{s} \subseteq \uprofile$, e.g., they are associated with one or more fairness agents.

As in a standard recommendation system we assume that we have (one or more) recommendation mechanisms that take a user profile $\uprofile$ and a (set of) items $\itm$ and produces a predicted rating $\prating \in \reals_+$. We will often refer to a recommendation list, $ \ell = \langle \{\itm_1, \prating_1\}, \ldots \{\itm_i, \prating_i\} \rangle $, which is generated for user $\uprofile$ by sorting according to $\prating$, i.e., $sort(\recfn_i(\uprofile, \items)) \rightarrow \ell$. Note that this produces a permutation (ranking) over the set of items for that user, i.e. a recommendation. As a practical matter, the recommendation results will almost always contain a subset of the total set of items, typically the head (prefix) of the permutation up to some cutoff number of items or score value. For ease of exposition we assume we are able to score all items in the database.

In the SCRUF-D architecture, fairness concerns map directly onto agents $\fagents = \{f_1, \ldots, f_i\}$. In order for the agents to be able to evaluate their particular concerns, they take account of the current state of the system and voice their evaluation of how fairly the overall system is currently operating, their compatibility for the current recommendation opportunity, and their preference for how to make the outcomes more fair. Hence, each fairness agent $i$ is described as a tuple, $f_i = \{m_i, c_i, \recfn_i\}$ consisting of a fairness metric, $m_i(\choicehist, \allochist) \rightarrow [0,1]$, that takes a choice history $\choicehist$ and allocation history $\allochist$ and produces a value in $[0,1]$ according to the agent's evaluation of how fair recommendations so far have been; a compatibility metric, $c_i(\uprofile) \rightarrow [0,1]$, that takes a particular user profile $\uprofile$ and produces a value in $[0,1]$ for how compatible fairness agent $i$ believes they are for user $\uprofile$; and a ranking function, $\recfn_i(\uprofile, v) \rightarrow \{\itm, \prating\}$, that gives the fairness agent preferences.

In the allocation phase (Figure \ref{fig:alloc}), we must allocate a set of fairness agents to a recommendation opportunity. Formally, this is an allocation function, $\alloc (\fagents, m_\fagents(\choicehist, \allochist), c_\fagents(\uprofile)) \rightarrow \beta \in \reals_{+}^{|\fagents|} $ that takes a set of fairness agents $\fagents$, the agents' fairness metric evaluations $m_\fagents(\choicehist, \allochist)$, and the agents' compatibility metric evaluations $c_\fagents(\uprofile)$ and maps to an agent allocation $\allocation$, where $\allocation$ is a probability distribution over the agents $\fagents$. The allocation function itself is allocating fairness agents to recommendation opportunities by considering both the fairness metric for each agent as well as each fairness agent's estimation of their compatibility.

The allocation function can take many forms, e.g., it could be a simple function of which every agent voices the most unfairness in the recent history \cite{sonboli2020and}, or it could be a more complex function from social choice theory such as the probabilistic serial mechanism \cite{bogomolnaia2001new} or other fair division or allocation mechanisms. Note here that the allocation mechanisms is directly comparing the agent valuations of both the current system fairness and compatibility. Hence, we are implicitly assuming that the agent fairness evaluations are comparable. While this is a somewhat strong assumption, it is less strong than assuming that fairness and other metrics, e.g., utility or revenue, are comparable as is common in the literature \cite{zhu2018fairness}. So, although we are assuming different normalized fairness values are comparable, we are only assuming that fairness is comparable with fairness, and not other aspects of the system. We explore options for the allocation function in our empirical experiments below. We track the outputs of this function as the allocation history, $\allochist = \langle \allocation^1, \ldots, \allocation^t \rangle$, an ordered list of agent allocations  $\allocation$ at time $t$. 

In the second phase of the system (Figure \ref{fig:choice}), we take the set of allocated agents and combine their preferences (and weights) with those of the current user $\omega$. To do this we define a choice function, $\choice(\ell, \beta, \ell_\fagents) \rightarrow \ell_\choice$, as a function from a recommendation list $\ell$, agent allocation $\allocation$, and fairness agent recommendation list(s) $\ell_\fagents$ to a combined list $\ell_\choice$. Each of the fairness agents is able to express their preferences over the set of items for a particular user, $\recfn_i(\uprofile, v) \rightarrow \{\itm, \prating\}$, and we take this set of lists,$\ell_\fagents =  \{\recfn_1(\uprofile, \items), \ldots, \recfn_i(\uprofile, \items)\}$, as input to the choice function that generates a final recommendation that is shown to the user, $\ell_\choice$. 

We again leave this choice function unspecified as this formulation provides a large design space: we could use a simple voting rule, a simple additive utility function or something much more complicated like rankings over the set of all rankings \cite{BCELP16a}. Note that the choice function can use the agent allocation $\beta$ as either a lottery to, e.g., select one agent to voice their fairness concerns, or as a weighting scheme. We investigate a range of choice functions in our experiments. In order for the fairness agents to be able to evaluate the status of the system we also track the choice history, $\choicehist = \langle \ell^t, \ell^t_\fagents, \ell^t_C \rangle$, as an ordered list of user recommendation list $\ell$, agent recommendation list(s) $\ell_\fagents$, and choice function output lists $\ell_\choice$, indexed by time step $t$.

\section{Design Considerations}
Within this framework there are a number of important design considerations to take into account for any particular instantiation of the SCRUF-D architecture. We have left many of the particular design choices open for future investigation. We allow for any type of recommendation algorithm; fairness agents may incorporate any type of compatibility function or  fairness evaluation function. Similarly, we do not constrain the allocation or choice mechanisms. With SCRUF-D, we are able to explore many definitions of fairness and recommendation together in a principled and uniform way. In this section, we discuss a few of the design parameters that may be explored in future work. 

\subsection{Agent Design}
We can expect that an agent associated with a fairness concern will typically have preferences that order items relative to a particular feature or features associated with that concern. Items more closely related to the sphere of concern will be ranked more highly and those unrelated, lower. However, this property means that agents associated with different concerns might have quite different rankings -- the gender parity concern will rank women's loans highly regardless of their geography, for example. Thus, we cannot assume consistency or single-peakedness across the different agents.

As noted above, agents may have preferences over disjoint sets of items or they may be constrained only to have preferences over the items produced by the recommender system for the given user. This second option corresponds to a commonly-used \textit{re-ranking} approach, where the personalization aspect of the system controls what items can be considered for recommendation and fairness considerations re-order the list \cite{ekstrand2022fairness}. 
If an agent can introduce any item into its preferences, then we may have the challenge in the choice phase of integrating items that are ranked by some agents but not others. Some practical work-arounds might include a constraint on the recommender system to always return a minimum number of items of interest to the allocated agents or a default score to assign to items not otherwise ranked.

Despite our fairness-oriented motivation, it should be clear that our architecture is sufficiently general that an agent could be designed that pushes the system to act in harmful and unfair ways rather than beneficial and fairness-enhancing ones. The system has no inherent representation of fairness and would not be able to detect such usage. Thus, the importance of the initial step of stakeholder consultation and the careful crafting of fairness concerns. Because fairness concerns are developed within a single organization and with beneficence in mind, we assume that we do not need to protect against adversarial behavior, such as collusion among agents or strategic manipulation of preferences. The fact that the agents are all ``on the same team'' allows us to avoid constraints and complexities that otherwise arise in multi-agent decision contexts.  

\subsection{Agent Efficacy}
The ability of an agent to address its associated fairness concern in non-deterministic. It is possible that the agent may be allocated to a particular user interaction, but its associated fairness metric may still fail to improve. One likely reason for this is the primacy of the personalization objective. Generally, we expect that the user's interests will have the greatest weight in the final recommendations delivered. Otherwise, the system might have unacceptably low accuracy, and fail in its primary information access objective. 

One design decision therefore is whether (and how) to track agent efficacy as part of the system history. If the agent's efficacy is generally low, then opportunities to which it is suited become particularly valuable; they are the rare situations in which this fairness goal can be addressed. Another aspect of efficacy is that relationships among item characteristics may mean that a given agent, while targeted to a specific fairness concern, might have the effect of enhancing multiple dimensions of fairness at once. Consider a situation in which geographic concerns and sectoral concerns intersect. Promoting an under-served region might also promote an under-served economic sector. Thus, the empirically-observed multidimensional impact of a fairness concern will need to be tracked to understand its efficacy.

Efficacy may also be a function of internal parameters of the agent itself. A separate learning mechanism could then be deployed to optimize these parameters on the basis of allocation, choice and user interaction outcomes.

\subsection{Mechanism Inputs}
Different SCRUF implementations may differ in what aspects of the context are known to the allocation and/or choice mechanisms. Our hope is that we can leverage social choice functions in order to limit the complexity of the information that must be passed to the allocation and/or choice mechanisms. However, if a sophisticated and dynamic representation of agent efficacy is required, it may be necessary to implement a bandit-type mechanism to explore the space of allocation probabilities and/or agent parameters as discussed above. Recent research on multidimensional bandit learning suggests possible approaches here \cite{mehrotra2020bandit}.

\subsection{Agent Priority}
As we have shown, agent priority in the allocation phase may be a function of user interests, considering different users as different opportunities to pursue fairness goals. It may also be a function of the history of prior allocations, or the state of the fairness concerns relative to some fairness metric we are trying to optimize. As the efficacy consideration would indicate, merely tracking allocation frequency is probably insufficient and it is necessary to tie agent priority to the state of fairness. Allocation priority is also tied to efficacy as noted above. It may be necessary to compute expected fairness impact across all dimensions in order to optimize the allocation.

We plan to leverage aspects of social choice theory to help ameliorate some of these issues. There is a significant body of research on allocation and fair division mechanisms that provide a range of desirable normative properties including envy-freeness \cite{cohler2011optimal}, e.g., the guarantee that one agent will not desire another agent's allocation, Pareto optimally, e.g., that agents receive an allocation that is highly desirable according to their compatibility evaluations \cite{bogomolnaia2001new}. An important and exciting direction for research is understanding what allocation properties can be guaranteed for the SCRUF-D architecture overall depending on the allocation mechanism selected \cite{BCELP16a}. 

We note that in most practical settings the personalization goal of the system will be most important and therefore the preference of this agent will have topmost priority. It is always allocated and is not part of the allocation mechanism. Thus, we cannot assume that the preference lists of the agents that are input to the choice system are anonymous, a common assumption in the social choice literature on voting \cite{BCELP16a}.

\subsection{Bossiness}
Depending on how the concept of agent / user compatibility is implemented, it may provide benefits to \textit{bossy} users, those with very narrow majoritarian interests that do not allow for the support of the system's fairness concerns. Those users get results that are maximally personalized and do not share in any of the potential accuracy losses associated with satisfying the system's fairness objectives. Other, more tolerant users, bear these costs. A system may wish to ensure that all users contribute, at some minimal level, to the fairness goals. In social choice theory, a mechanism is said to be non-bossy if an agent cannot change the allocation without changing the allocation that they receive by modifying their preferences \cite{papai2000strategyproof}. Some preliminary discussions of this problem specifically for fairness-aware recommendation appear in \cite{farastu2022pays}. 

\subsection{Fairness Types}
We concentrate in this paper and our work with Kiva generally on provider-side group fairness, that is characteristics of loans where protected groups can be distinguished. However, it is also possible to use the framework for other fairness requirements. On the provider side, an individual fairness concern is one that tracks individual item exposure as opposed to the group as a whole. It would have a more complex means of assessing preference over items and of assessing fairness state, but still fits within the framework.

Consumer-side fairness can also be implemented through use of the compatibility function associated with  each agent. For example, the example of assigning risk appropriately based on user risk tolerance becomes a matter of having a risk reduction agent that reports higher compatibility for users with lower risk tolerance.

\section{Experimental Methodology}

As an initial examination of the properties of the SCRUF-D architecture, we conducted a series of experiments with real and simulated data, run on a Python implementation of the SCRUF-D architecture. See associated GitHub repository for the source code.\footnote{https://github.com/that-recsys-lab/scruf\_d}. Configuration files, data and Jupyter notebooks for producing the experiments and visualizations are found in a separate repository\footnote{https://github.com/that-recsys-lab/scruf\_tors\_2023}. 

\subsection{Data sets}

\subsubsection{Microlending Data}
We used the Microlending 2017 dataset \cite{sonboli2022micro}, which contains anonymized lending transactions from Kiva.org. The dataset has 2,673 pseudo-items, 4,005 lenders and 110,371 ratings / lending actions. See \cite{sonboli2020opportunistic} and \cite{sonboli2022micro} for a complete description of the data set. 

We considered two loan feature categories, loan size and country, as protected features. Prior work \cite{sonboli2020opportunistic} identified loan size as a dimension along which fairness in lending may need to be sought. About 4\% of loans had this feature and were considered protected items. We set the fairness target to be 20\%. For the second protected feature, we followed \citet{sonboli2020opportunistic} in identifying the 16 countries whose loans have the lowest rates of funding and labeled these as the protected group for the purposes of geographic fairness. We set the fairness target to be 30\%. Compatibility scores were defined using the entropy of a user's ratings versus the protected status of funded loans using the method in \cite{sonboli2020opportunistic}. 

\subsubsection{MovieLens data}

We also used the MovieLens 1M dataset \cite{movielens}, which contains user ratings for movies. The dataset has 3,900 movies, 6,040 users, and approximately 1 million ratings. We selected movies with female writers and directors as one protected category and movies with non-English scripts as the other. We set the fairness targets for these to be 12\% and 28\%, respectively, which mirrors their prelevance in the item catalog.

\subsubsection{Synthetic data}

The purpose of synthetic data in our simulations is to supply realistic recommender system output as input to the SCRUF-D reranker, allowing experimental control of the number of sensitive features, the prevalence of sensitive features among items, and the differing receptiveness of users towards those features.

We create synthetic data via latent factor generation. That is, we create matrices of latent factors similar to those that would be created through factorization and then generate sample ratings from these matrices. Let $\hat{U}$ and  $\hat{V}$ be the user and item latent factor matrices with $k$ latent factors. We designate the first $k_s$ of the latent factors as corresponding to protected features of items, and the remaining $k - k_s$ factors correspond to other aspects of the items.

As a first step, we generate a vector of real-valued propensities for each user $\Phi_i = \langle\phi_1, ..., \phi_{k_s}\rangle$ corresponding to the sensitive features plus additional values for each of the non-sensitive features, drawn from an experimenter-specified normal distribution. Thus, it is possible to adjust the preferences of the user population regarding different sensitive features. The propensities associated with a sensitive feature also represent the user's compatibility with the respective fairness agent, a value which in a non-synthetic case is derived from the pre-existing user profile as in \cite{sonboli2020opportunistic}. 

From $\Phi_i$, we perform an additional generation step to draw a latent factor vector $U_i$ from a normal distribution centered on the propensities. This two-step process avoids having the latent factors tied exactly to the user propensities, which would otherwise make the compatibility of users with agents highly deterministic. 

The profiles for items are generated in a similar way except that items have a binary association with their associated sensitive features and so the experimenter input consists of parameters for a multi-variate Bernoulli distribution. Each item's propensity is generated as a binary vector $\Phi_j$ using these probabilities. As with users, there is a second step of latent factor generation, in which the elements of an item's latent feature vector $V_j$ is drawn from a normal distribution centered on the item's (binary) propensity for that feature. This two-step procedure allows us to identify an item as possessing a particular feature (particularly protected ones) without the latent factor encoding this exactly. 

After $\hat{U}$ and $\hat{V}$ have been generated, we then select $m$ items at random for each user $i$ and compute the product of the $\hat{U}_i$ and $\hat{V}_j$ as the synthetic rating for each user $i$, item $j$ pair. To simulate the bias for which a fairness solution is sought, we impose a rating penalty $\gamma$ on ratings generated for items with sensitive attributes. We sort these values and select the top $m'$ as the recommender system output. The sorting / filtering process ensures that the output is biased towards more highly-rated items, which is what one would expect in recommender system output.

For the experiments in this paper, we generated 1,500 users (500 users with a high propensity towards the first factor, 500 users with a high propensity for the second protected factor and 500 users with an average propensity for both) and 1,000 items. For each user, we generated 200 sample ratings and used the top 50 as the recommendation lists. Item propensities were set to 0.1, 0.3 for the first two sensitive factors and the other values were randomly set. The standard deviation of the factors was 1.0. Corresponding user propensities for the features were $\mu = 0.5, \sigma = 0.05$ for the both protected factors in the case of the average propensity batch of users, $\mu = 0.1, \sigma = 0.1$ for the low propensity factor and $\mu = 0.9, \sigma = 0.1$ for the high propensity factor in the other two options. The generation parameters were based on proportions seen in real-world datasets including the Microlending dataset described above. The bias penalty $\gamma$ was set to 3.0.

To explore the dynamic response of the system, we created an artificial ordering of the synthetic users with three segments $<A, B, C>$, each arriving in sequence. We placed the synthetic users with high compatibility to Agent 2 and low compatibility with Agent 1 in segment $A$, then reversed this affinity in segment $B$. Segment $C$ contained users without high compatibility with either agent. This data is referred to as the $Synthetic$ dataset in the experiments.

 \subsection{Fairness metrics}

The agent-specific fairness metric allows each agent to calculate their current state of fairness given the user interactions that have occurred within the evaluation window. As we have noted above, SCRUF allows for a wide variety of metrics and makes no assumptions that agents have metrics with similar logic. For the experiments in this study, we have chosen to have uniform fairness metric across agents to more easily assess the impact of varying other platform parameters including the allocation and choice mechanisms. 

For each agent defined on fairness concern relative to each dataset, we assign a target exposure value as noted above. That is, an agent with a 20\% target exposure will return a value of 1.0 (perfect fairness) if, across all of the recommendation lists in the evaluation window, there is at least an average of 20\% of items with its associated protected feature. More items do not result in a higher score, but fewer items would yield a value linearly-scaled towards zero, as the value when no protected items have been recommended. We have set the targets artificially high to investigate how the system respond to this pressure to include more protected items.

\subsection{Mechanisms}

As noted above, there is a wide variety of different allocation mechanisms that can be studied. For our purposes in this paper, we are exploring mechanisms with widely differing logics to understand the implication of these choices for recommendation outcomes.

\begin{itemize}
    \item \textbf{Least Fair:} The fairness agent with the lowest fairness score $m_i$ is chosen. This simple and commonly-used rule ensures that low fairness agents get attention, but it does not take into account the compatibility between a user and an item and so may cause more accuracy loss than others. 

    \item \textbf{Lottery:} A lottery is constructed with probabilities proportional to the product of agent unfairness and compatibility: $p(f_i) \propto (1 - m_i) * c_i$, normalized to sum to 1. A single agent is chosen by drawing from this lottery. 

   \item \textbf{Weighted:} All agents are allocated to every recommendation opportunity but their weight is determined by the product of their unfairness and compatibility, similar to the lottery probabilities above: $\beta_i \propto (1 - m_i) * c_i$, normalized.
\end{itemize}

\subsection{Choice Mechanisms (Voting Rules)}
We examine four different choice mechanisms. In computational social choice, choice mechanisms are classically understood as integrating the preferences of multiple agents together to form a single societal preference \cite{BCELP16a}.\footnote{Our setting differs from classical social choice in that voting is not anonymous (the recommender system plays a different role from the other agents) and weights and scores are typically employed. Typically, rankings are preferred because they do not require agent utility to be known or knowable.} 
\\
\textbf{Rescore:} The simplest mechanism is one in which each agent contributes a weighted score for each item and these scores are summed to determine the rank of items. Each fairness agent has a fixed score increment $\delta$ that is added to all protected items, weighted by its allocation in the previous phase. This is combined with the scores computed by the recommendation algorithm. \\
\textbf{Borda:} Under the Borda mechanism \cite{Zwicker:Voting}, ranks are associated with scores and the original scores used to compute those ranks are ignored. The ranks across agents are summed and the result determines the final ranking. \\
\textbf{Copeland:} The Copeland mechanism calculates a win-loss record for each item considering all item-item pairs in a graph induced by the preferences. Item $i$ scores one point over item $j$ if the majority of allocated agents prefer $i$ to $j$. We then sum these pairwise match-ups for each item $i$ and order the list of items using these scores \cite{sep-voting-methods}. \\
\textbf{Ranked Pairs:} The Ranked Pairs voting rule \cite{tideman1987independence} computes the pairwise majority graph as described for Copeland but orders the resulting ranking by how much a particular item wins by, selecting these in order to create a complete ranking, skipping a pair if and only if it would induce a cycle in the aggregate ranking. 

Each of these choice mechanisms implements a fundamentally different logic for aggregating preferences: score-based, ordinal-based, consistency-based and pairwise-preference \cite{Zwicker:Voting}. As we show in our results, choice mechanisms yield quite different accuracy / fairness tradeoffs. 

While the agent weights in these mechanisms fixed by the allocation mechanism (and normalized to 1), the recommender weight is a parameter, which determines how much the recommender systems results are emphasized relative to the fairness agents. Under different conditions and mechanisms, different recommender weights may be optimal. We refer to this weight throughout as $\lambda$.

\subsection{Baseline algorithms}

As noted above, there are very few recommendation algorithms that allow for dynamic reranking with multiple fairness concerns at once. The Multinomial FA*IR method described in \cite{zehlike2022fair} is not practical for recommendation because of its time complexity. For these experiments, we use OFair \cite{sonboli2020opportunistic} and MultiFR \cite{wu2022multi}.

OFair is a multi-group fairness-aware recommendation reranking method based on the technique of maximum marginal relevance (MMR) \cite{carbonell1998use} for diversity enhancement in reranking. OFair seeks to enhance fairness for multiple groups by treating each sensitive aspect as a dimension of diversity and greedily building a recommendation list by adding items that enhance its diversity in this respect. OFair adds an additional consideration of personalization by weighting different features according to user compatibility, building on the work of \cite{liu2018personalizing} using profile entropy relative to the sensitive features. OFair also has a $\lambda$ parameter, controlling how much the recommender systems scores are weighted in the reranking process.

Multi-FR is described as a ``multi-objective optimization framework for fairness-aware recommendation'' \cite{wu2022multi}.  Multi-FR models fairness constraints using a smoothed stochastic ranking policy and optimizes for fairness constraints using multiple gradient descent. The method finds multiple solutions along a theoretical Pareto frontier and chooses the best solution using a least misery strategy. Note that Multi-FR uses a batch-oriented strategy, attempting to address the fairness concerns over the entire set of recommendations at once. Still, Multi-FR is one of the few existing algorithms for fairness-aware recommendation that supports multiple fairness concerns, including provider-side and consumer-side constraints. 

One key limitations of Multi-FR is that it represents provider-side fairness only in terms of mutually-exclusive provider groups. SCRUF-D has no such limitation and supports intersecting fairness concerns. Because of this difference, in our experiments, we had to create a cross-product of all possible combinations of protected features in order to capture the multiple features in our datasets. Also, Multi-FR only supports a single type of fairness objective on the provider side: minimizing the difference between actual and ideal exposure of item categories. For the experiments below, target exposures were set as follows: Microlending: loan size/0.20, country/0.30; both features/0.026; MovieLens: women writer or director/0.09, non-English/0.25, both features/0.02. Multi-FR has its own method of balancing accuracy and fairness and so does not have a parameter controlling the balance between fairness and accuracy. 

Batch-oriented processing as found in Multi-FR is a fairly common approach for fairness-aware recommendation \cite{ekstrand2022fairness} and it is true that recommendation results are often cached, so processing many users at once is a practical approach. However, it should be noted that a batch approach to fairness-aware recommendation does not guarantee fairness in the recommendations that are delivered. The system can guarantee that a good fairness/accuracy tradeoff is found across the recommendation lists that are processed in a given batch, but, these may not in fact be the recommendations that are delivered to users over any interval. The users compatible with a particular protected group (for example, those interested in foreign language movies) may not happen to show up very often. So, the careful balance between criteria achieved within the batch process may not be realized when the recommendations are delivered in practice: the recommendations which are fairness-enhancing may sit in the cache and never be output. One reason to prefer an on-demand approach to fairness enhancement is that it is responsive to fairness outcomes in the moment. Still, we have included Multi-FR as a comparator to indicate what batch-oriented algorithm can achieve by considering all the recommendations at once.

\subsection{Evaluation}

We evaluate ranking accuracy using normalized discounted cumulative gain (nDCG) and fairness using our (normalized) fair exposure relative to the target proportion set for each protected group. Note that the fairness metric as computed by each agent is different from the overall fairness computed over the experiment, for the simple reason that agents only look back over a fixed window in computing their fairness at each time point. We will use the notation $\bar{m_i}$ to refer to the global fairness for agent $i$.

To derive a single score representing both the combined fairness of both agents and the disparity between them, we use the $L_{1/2}$-norm, which for our purposes is defined as

\begin{equation}
L_{1/2} = \frac{1}{4}(\sum_{\forall i}{\sqrt{\bar{m}_i}})^2    
\end{equation}

The factor of $1/4$ is used to give the resulting value the same scale as the original fairness scores. If we consider a fixed average $\bar{m}^*$ across all the agents, the $L_{1/2}$ norm is maximized (and equal to $\bar{m}^^*$) if all of the fairness values are the same $\bar{m}_i = \bar{m}^*$. A mix of lower and higher values with the same average will give a lower result.

A dynamic way to look at local fairness is to consider \textit{fairness regret} over the course of the experiment. At each time step, we calculate $1 - m_i$, that is the difference from perfect fairness as the agent defines it, and then sum these values over the course of the simulation. This is similar to the notion of regret in reinforcement learning but using fairness instead of utility. Fairness regret $G_i$ for agent $i$ is defined as:

\begin{equation}
    G_i(s) = \sum_{t=0}^{s}{1 - m_i(\choicehist_t, \allochist_t)}
\end{equation}

\section{Results}

For the two datasets, we present results showing the results of adjusting $\lambda$, the weight associated with the recommender agent. Lower $\lambda$ values put more weight on the reranking mechanisms. We then select a single $\lambda$ value for further analysis of each combination of mechanisms.

\subsection{MovieLens dataset}

\begin{figure}[tbh]
    \centering
    \includegraphics[width=0.75\textwidth]{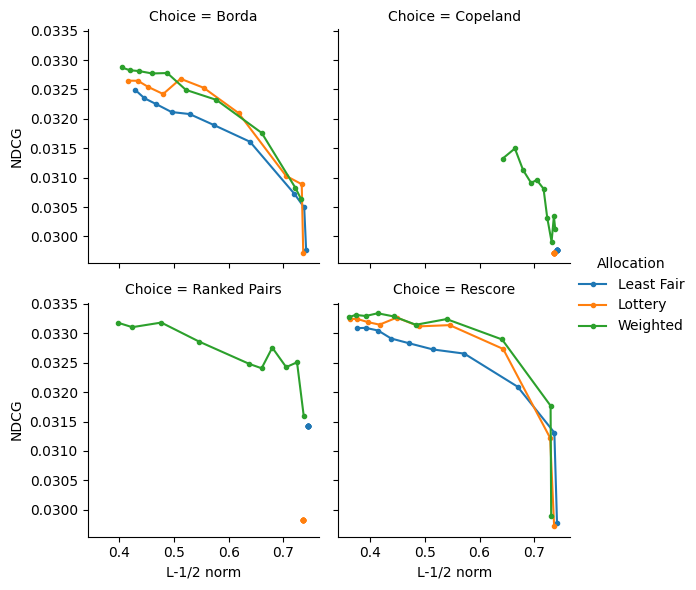}
    \caption{Accuracy vs fairness for the MovieLens dataset at different values of $\lambda$.}
    \label{fig:movie-lambda-results}
\end{figure}

Figure~\ref{fig:movie-lambda-results} shows the results for the MovieLens data organized by choice mechanism and showing the results of adjusting $\lambda$, the weight associated with the recommender system. In general, as one might expect, as the weight decreases, accuracy drops and fairness increases. We note that the pair-wise Copeland and Ranked Pairs mechanisms prove difficult to tune when there is only a single agent being allocated. There is only a single value for all cases when $\lambda < 1.0$. This is because these mechanisms depend only on rankings and so as long as the recommender outweighs the allocated agent, it wins all of the pair-wise comparisons and when it does not, it loses all of them. 

As was also seen in \cite{burke2018balanced}, there exist some ``win-win'' regions with the Rescore mechanisms in conjunction with Weighted allocation. For small decreases in $\lambda$, we see both fairness and accuracy increase at the same time, indicating that some reranking is actually beneficial to accuracy, even as measured in this off-line setting. In addition, the shallow slope of some of these curves suggests that implementers can increase fairness quite significantly over the baseline without too much loss of ranking accuracy. The other mechanisms have a steeper accuracy loss.

For the remainder of this discussion, we select the $\lambda$ values where accuracy loss is less than or equal to 5\% and consider what is the best fairness that can be achieved within this constraint. We do not have the ability to tune Multi-FR because of its design, which does Pareto optimization internally, so we report the results from this algorithm as designed.

\begin{table*}[tbh]
{\small
\begin{tabular}{|r|r|r|r|r|r|r|r|}
\hline
Allocation & Choice & $\lambda$ & nDCG & $\bar{m}_1$ & $\bar{m}_2$ & $L_{1/2}$ & Avg  \\
\hline
--- & --- & 1.0 & 0.0331 & 0.5116 & 0.1103 & 0.2742 & 0.3109 \\
\hline
--- & *OFair & 1.0 & 0.0261 & 0.8035 & 0.2818 & 0.5092 & 0.5426 \\
--- & Multi-FR & N/A & 0.0076 & 1.082 & 1.721 & 1.402 & 1.383 \\
\hline
Least Fair & Borda & 0.41 & 0.0316 & 1.0950 & 0.3051 & 0.6390 & 0.7000 \\
 & *Copeland & 0.11 & 0.0298 & 1.2601 & 0.3594 & 0.7414 & 0.8097 \\
 & Ranked Pairs & 0.11 & 0.0314 & 1.2043 & 0.3963 & 0.7456 & 0.8003 \\
 & Rescore & 0.31 & 0.0321 & 1.1532 & 0.3164 & 0.6694 & 0.7348 \\
\hline
Lottery & Borda & 0.41 & 0.0321 & 1.0528 & 0.2985 & 0.6181 & 0.6757 \\
 & *Copeland & 0.11 & 0.0297 & 1.2461 & 0.3596 & 0.7361 & 0.8028 \\
 & *Ranked Pairs & 0.11 & 0.0298 & 1.2439 & 0.3591 & 0.7350 & 0.8015 \\
 & Rescore & 0.31 & 0.0327 & 1.0991 & 0.3086 & 0.6431 & 0.7039 \\
\hline
Weighted & Borda & 0.31 & 0.0318 & 1.0483 & 0.3617 & 0.6604 & 0.7050 \\
 & Copeland & 0.91 & 0.0315 & 1.0053 & 0.3940 & 0.6645 & 0.6997 \\
 & Ranked Pairs & 0.11 & 0.0316 & 1.1853 & 0.3947 & 0.7370 & 0.7900 \\
 & Rescore & 0.21 & 0.0318 & 1.2393 & 0.3542 & 0.7296 & 0.7967 \\
\hline
\end{tabular}
}
\caption{Accuracy and fairness results for the MovieLens data. Results were chosen to be the greatest $L_{1/2}$ fairness with nDCG loss no greater than 5\% over baseline, except for the mechanisms indicated by * which were unable to hit this target at any setting.}
\label{tab:movielens-results}
\end{table*}

Table~\ref{tab:movielens-results} includes all of the findings across the different mechanisms. We include both the $L_{1/2}$ norm and the average in the table. Where these are close in value, the agents are getting similar fairness outcomes. We can see that the \textit{women writers and directors} fairness target is quite a bit more difficult to hit than the \textit{non-English} target. There is a large difference already in the unreranked baseline and this carries through to the rerankers. Many of them are able to achieve and exceed the fairness target for the non-English feature, but none do better than 0.4 for the other feature. One exception is Multi-FR, which overshoots the fairness targets and ends up with very low accuracy. Both Least Fair and Weighted in conjunction with the Ranked Pairs mechanism do well in this respect. However, these mechanisms are not the best with respect to maintaining accuracy.

\begin{figure}
    \centering
    \includegraphics[width=0.75\textwidth]{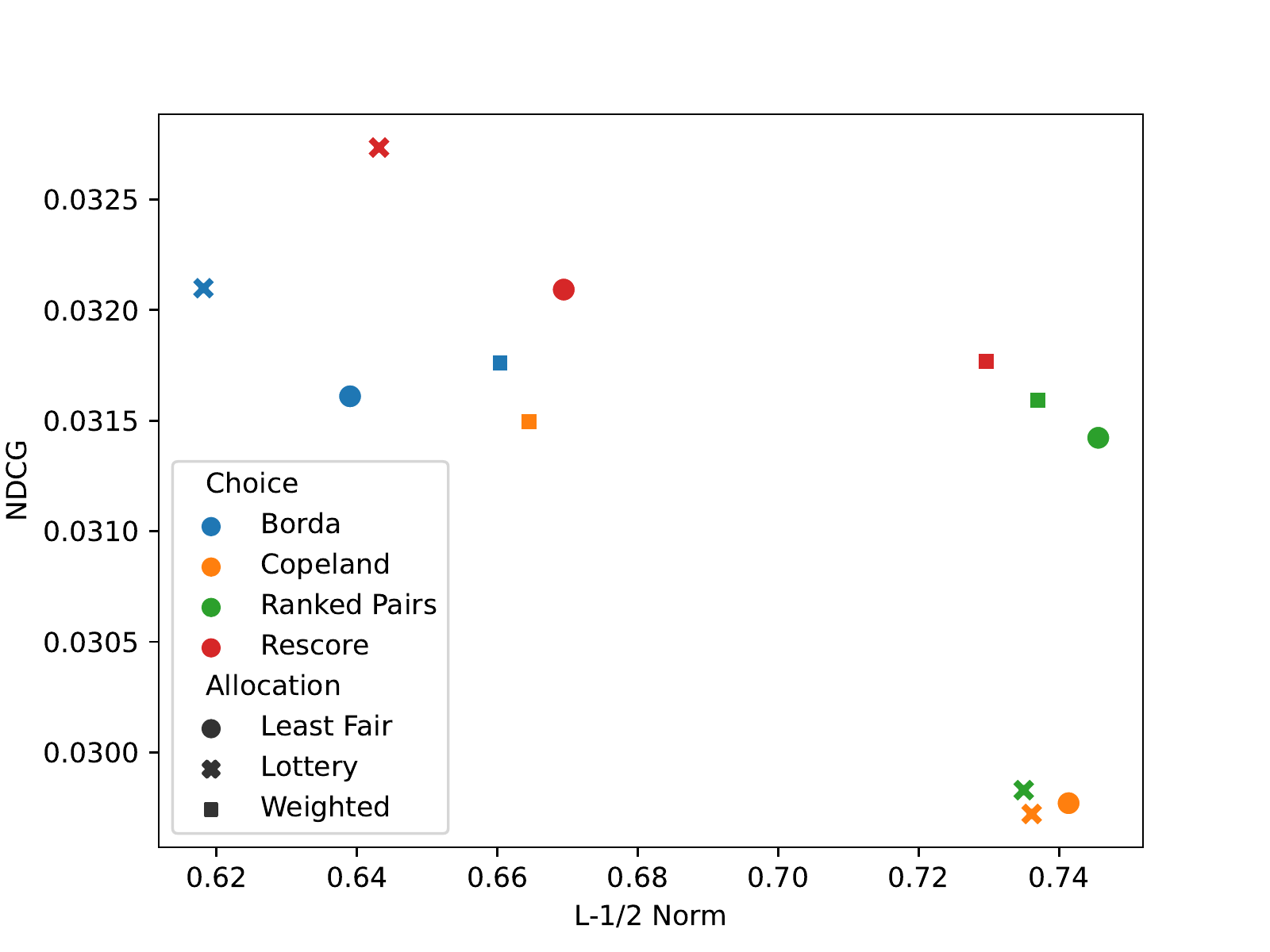}
    \caption{Comparison of mechanisms on MovieLens data. OFair is omitted as it is far off the chart to the lower left. Multi-FR is far below in terms of accuracy.}
    \label{fig:movie-full-results}
\end{figure}

Figure~\ref{fig:movie-full-results} shows the accuracy vs. fairness results for all of the mechanisms and the baseline algorithms. We see some clustering by choice mechanism, except for the two pair-wise algorithms, Copeland and Ranked Pairs. For these algorithms, the Lottery mechanism yields very poor ranking accuracy. For Copeland, this is also true of the Least Fair mechanism. The reason is that the Weighted mechanism is so different here is that it is bringing multiple agents to the choice mechanism in these cases. (The others only promote a single agent to the choice phase.) With these multiple agents in the mix, the recommender is dominant and so the agents are less effective at increasing fairness.

Except for the Rescore mechanism, the Lottery allocation data points are all dominated by other points along the Pareto frontier, which would seem to indicate a disadvantage to allocating only a single agent in the allocation phase. Theoretically, Lottery and Weighted allocations are the same in expectation since the lottery is drawn from the same numerical distribution. However, that is not borne out in the results here. In the case of the Rescore mechanism, these two allocation mechanisms represent very different positions in the tradeoff space. 

Overall, the Rescore and Ranked Pairs mechanisms occupy the dominant positions: Rescore at higher accuracy levels and Ranked Pairs at higher fairness levels. Somewhat surprisingly, the Least Fair allocation stakes out two spots on the Pareto frontier, even though it ignores user compatibility. It is generally lower in accuracy than Lottery or Weighted mechanisms when the same choice mechanism is applied although there are exceptions in the experiments.

Both OFair and Multi-FR are dominated by the SCRUF mechanisms at different points. In the case of Multi-FR, it appears to have inherent limitations on how much accuracy loss it is willing to entertain in order to increase fairness. We also note that its reranking decisions are made off-line in a batch mechanism and so it is not able to respond dynamically to fairness issues in the moment. OFair also has lower fairness and accuracy. Like the Weighted mechanism, it is trying to address all of the fairness concerns at once in each recommendation list.

\subsection{Microlending dataset}

\begin{figure}
    \centering
    \includegraphics[width=0.75\textwidth]{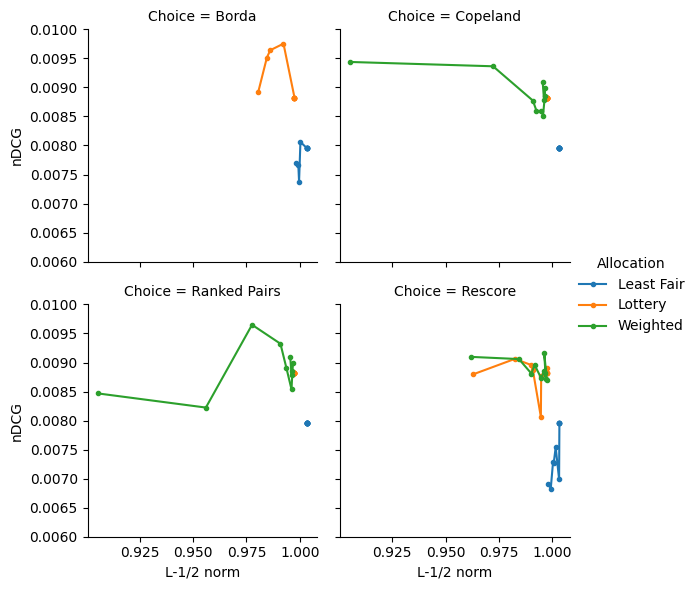}
    \caption{Accuracy vs fairness for the Microlending dataset at different values of $\lambda$. Value for Weighted + Borda had much lower fairness and were omitted.}
    \label{fig:kiva-lambda-results}
\end{figure}

Figure~\ref{fig:kiva-lambda-results} shows the results for the Microlending dataset organized by choice mechanism adjusting $\lambda$, the weight associated with the recommender system. The Microlending dataset turns out to be quite different from MovieLens for several reasons. One is that the fairness targets are much easier to achieve. As can be seen in Figure~\ref{fig:kiva-lambda-results}, almost all of the mechanisms are able to achieve fairness very close to 1.0. We also see fairly noisy behavior across the different $\lambda$ values, rather than the smoother accuracy / fairness tradeoff seen in the MovieLens data. We believe that this is a side-effect of the easier fairness target: there is not much trading-off that the system needs to perform.

\begin{table*}[tbh]
{\small
\begin{tabular}{|r|r|r|r|r|r|r|r|}
\hline
Allocation & Choice & $\lambda$ & nDCG & $\bar{m}_1$ & $\bar{m}_2$ & $L_{1/2}$ & Avg \\
\hline
None & None & 1.0 & 0.0074 & 0.6564 & 0.0397 & 0.3481 & 0.2547\\
\hline
None & OFair & 1.0 & 0.0080 & 0.6616 & 0.4539 & 0.5529 & 0.5578 \\
\hline
None & Multi-FR & N/A & 0.0262 & 0.4472 & 0.6205 & 0.5338 & 0.5303 \\
\hline
Least Fair & Borda & 0.61 & 0.0080 & 1.0077 & 0.9991 & 1.0034 & 1.0034 \\
 & Copeland & 0.71 & 0.0080 & 1.0077 & 0.9991 & 1.0034 & 1.0034 \\
 & Ranked Pairs & 0.71 & 0.0080 & 1.0077 & 0.9991 & 1.0034 & 1.0034 \\
 & Rescore & 0.41 & 0.0080 & 1.0077 & 0.9991 & 1.0034 & 1.0034 \\
\hline
Lottery & Borda & 0.61 & 0.0088 & 1.0011 & 0.9940 & 0.9975 & 0.9975 \\
 & Copeland & 0.71 & 0.0088 & 1.0011 & 0.9940 & 0.9975 & 0.9975 \\
 & Ranked Pairs & 0.51 & 0.0088 & 1.0011 & 0.9940 & 0.9975 & 0.9975 \\
 & Rescore & 0.41 & 0.0089 & 1.0015 & 0.9940 & 0.9977 & 0.9978 \\
\hline
Weighted & Borda & 0.61 & 0.0082 & 0.6548 & 0.5054 & 0.5777 & 0.5801 \\
 & Copeland & 0.31 & 0.0090 & 0.9995 & 0.9938 & 0.9966 & 0.9966 \\
 & Ranked Pairs & 0.31 & 0.0090 & 0.9995 & 0.9938 & 0.9966 & 0.9966 \\
 & Rescore & 0.41 & 0.0087 & 0.9995 & 0.9955 & 0.9975 & 0.9975 \\
\hline
\end{tabular}
}
\caption{Accuracy and fairness results for the Microlending dataset.  Results were chosen to be the best tradeoff for each respective mechanism}
\label{tab:kiva_full_results}
\end{table*}

Table~\ref{tab:kiva_full_results} shows the complete results and we note that the mechanisms are finding identical solutions in many cases, with the same fairness and accuracy values. This suggests that the original results contain only a limited number of protected items in each list and so there is only so much room for the reranker to alter the results. We believe that part of the reason for this effect is our choice of biased matrix factorization as our base recommendation algorithm. This algorithm is known to suffer from popularity bias and be limited in the range of items that it recommends \cite{jannach2015recommenders}. In our future work, we will examine alternate base algorithms with better diversity.

\begin{figure}
    \centering
    \includegraphics[width=0.75\textwidth]{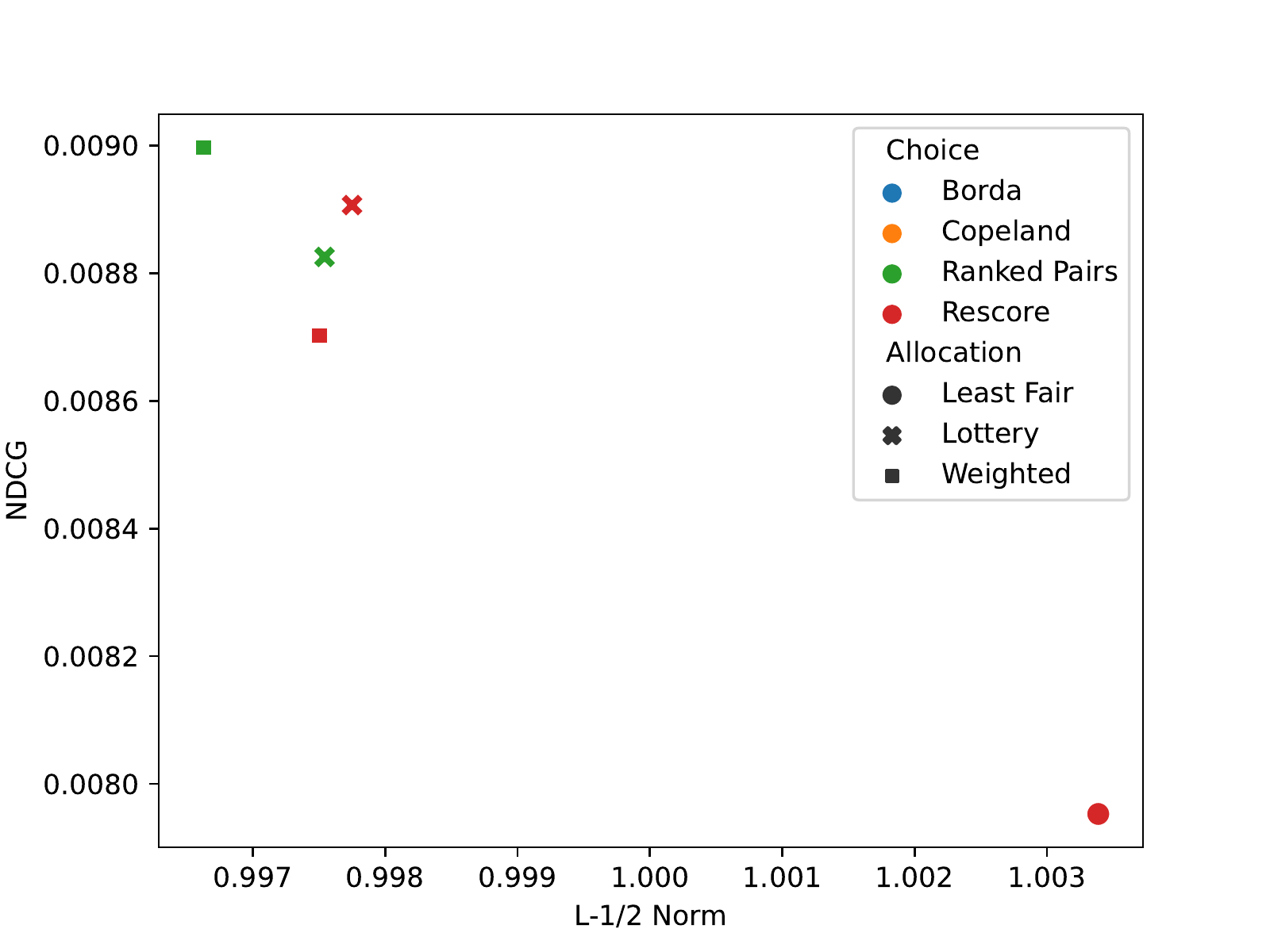}
    \caption{Comparison of mechanisms on Microlending data. OFair is omitted as it is far off the chart to the lower left. Multi-FR has higher NDCG but is also omitted due to its far lower fairness.}
    \label{fig:kiva-full-results}
\end{figure}

The results for the Microlending data are summarized in the scatterplot in Figure~\ref{fig:kiva-full-results}. There only five points representing the 14 experimental conditions. OFair and Multi-FR are omitted because they are far from the optimal tradeoff region (although we know from the table that Multi-FR has quite good accuracy). All four of the Least Fair conditions have the same values at the lower right, as do many of the Lottery conditions, except for Rescore. 

With this limited data, it is hard to draw too many conclusions. Unlike in the MovieLens case, Least Fair occupies only the most extreme lower accuracy condition in this data. Rescore still seems to be a good strategy and Ranked Pairs with the Weighted allocation is still on the Pareto frontier as it was in the MovieLens case.

\subsection{Fairness dynamics}

The compatibility of users with diverse sensitive features turns out to be highly correlated in our real-world datasets, which is perhaps not surprising, but it makes it difficult to explore dynamic scenarios simulating the arrival of disparate types of users at different times. For this reason, we used the $Synthetic$ dataset described above in simulated experiments of user arrivals to examine how the balance between agents is achieved over time. With this synthetic data set, we do not have ground truth user preferences and so we evaluate recommendation accuracy only relative to the original rankings in the simulated data.  
 
{\setlength\textfloatsep{10pt}
\begin{figure}[tb]
   \centering
    \begin{subfigure}[b]{0.45\textwidth}
        \centering
        \includegraphics[scale=0.25]{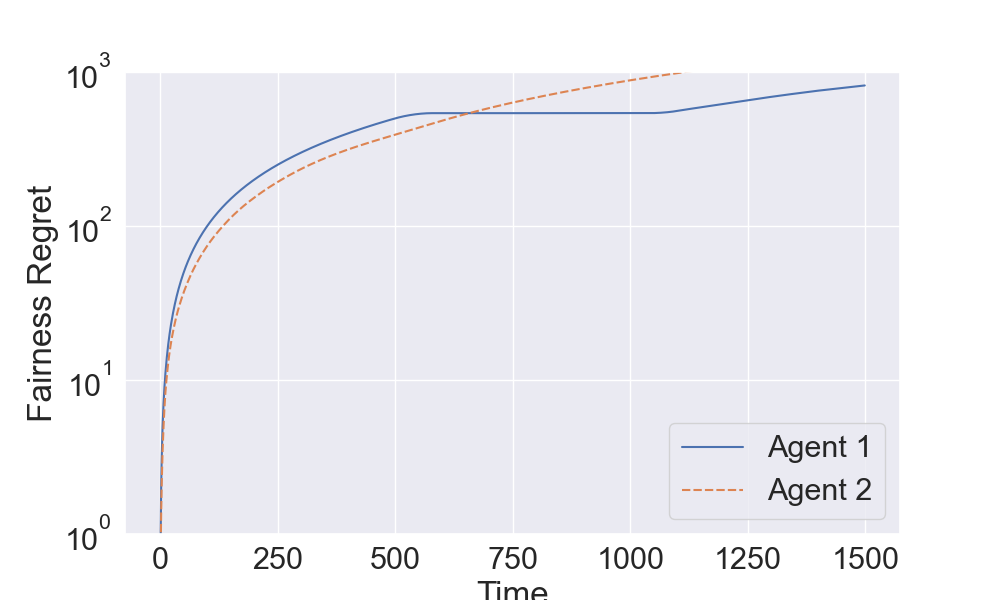}
        \caption{Baseline}
        \label{fig:baseline-regret}
    \end{subfigure}
    \hfill
    \begin{subfigure}[b]{0.45\textwidth}
        \centering
        \includegraphics[scale=0.25]{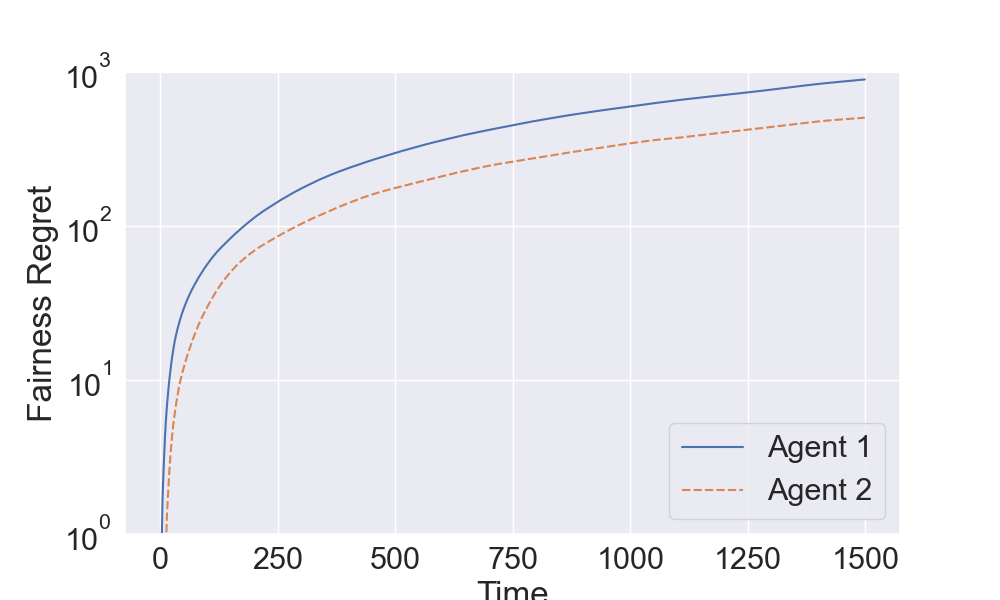}
        \caption{OFair}
        \label{fig:ofair-regret}
    \end{subfigure} \\
    \begin{subfigure}[b]{0.45\textwidth}
        \centering
        \includegraphics[scale=0.25]{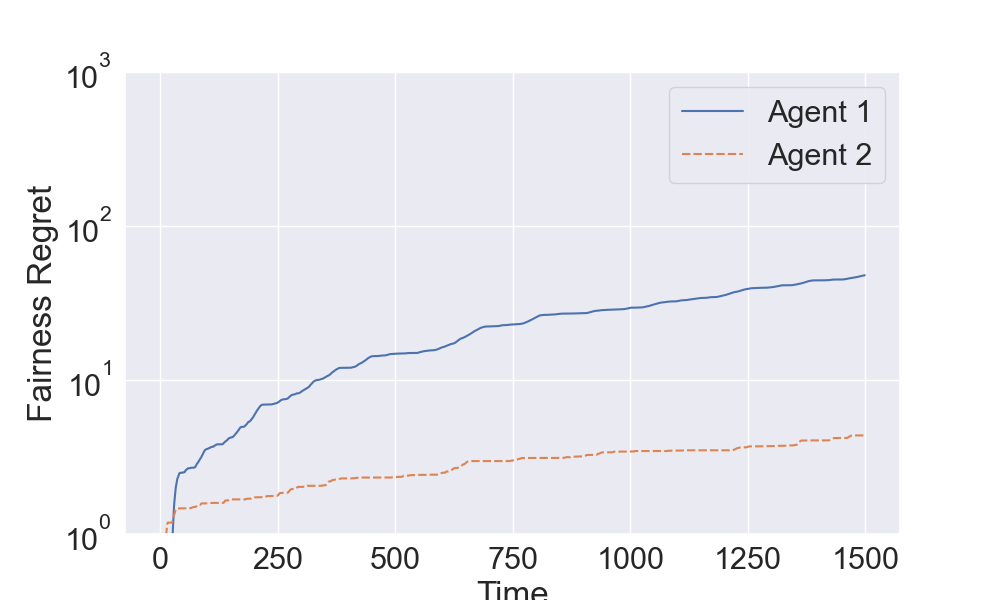}
        \caption{Lottery}
        \label{fig:lottery-regret}
    \end{subfigure}
    \hfill
    \begin{subfigure}[b]{0.45\textwidth}
        \centering
        \includegraphics[scale=0.25]{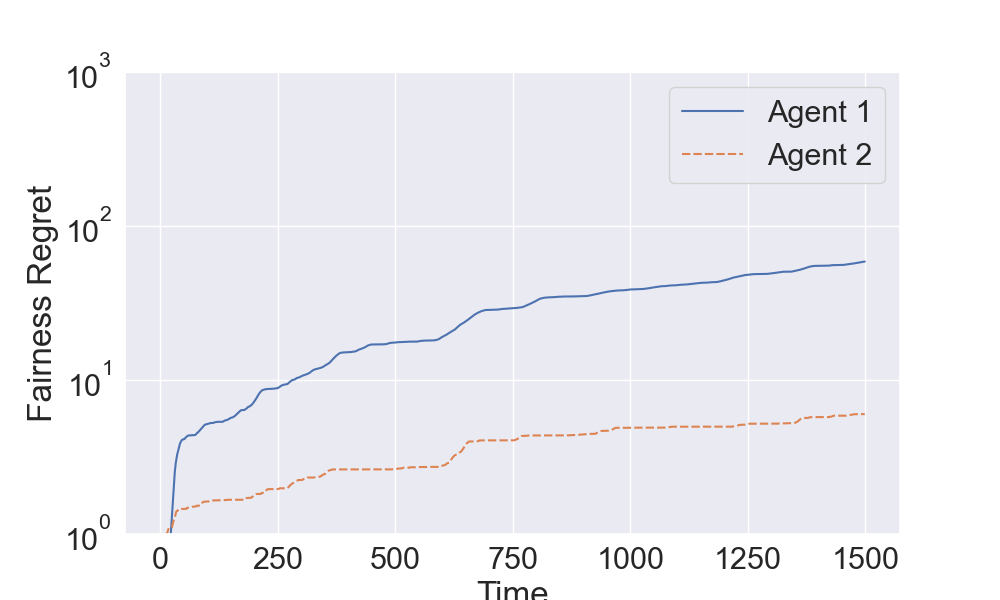}
        \caption{Weighted}
        \label{fig:weighted-regret}
    \end{subfigure} \\
    \begin{subfigure}[b]{0.45\textwidth}
        \centering
        \includegraphics[scale=0.25]{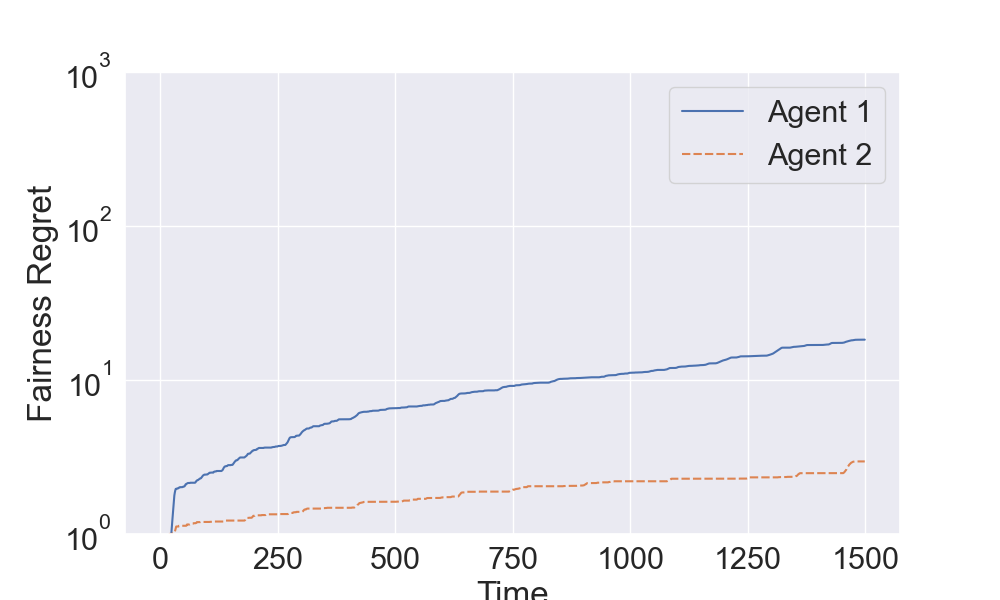}
        \caption{Least Fair}
        \label{fig:least-regret}
    \end{subfigure} 
    \caption{Cumulative fairness regret for fairness agents with different allocation mechanisms}
    \label{fig:dynamic-regret}
\end{figure}
}

In Figure~\ref{fig:dynamic-regret}, we look closely at the cumulative fairness regret for the different allocation mechanisms. We keep the choice mechanism fixed (Borda) to isolate the impact of allocation on agent outcomes. Recall that the arrival of users is segmented so that users compatible with Agent 2 arrive as the first 500, followed by another 500 compatible with Agent 1, and then a third 500 user segment without strong compatibility to either agent. Note that the y-axis on these plots has a log scale. 

For the Baseline algorithm, without reranking, the impact of the different segments can be see in the flattening of the Agent 1 curve for users 500-1000. These users are compatible with Agent 1 and already have some of these protected items in their recommendation lists. The regret ends up quite high for both agents.

The OFair algorithm does not fare much better. It is trying to satisfy all of the fairness constraints at once. While its fairness regret is lower, especially for Agent 2, it is still quite high. The other allocation mechanisms fare much better, maintaining 10x or greater improvement in regret over the course of the experiment. The Lottery and Weighted mechanisms are quite similar to each other, with the Weighted mechanism doing slightly better for Agent 2. By not trying (as hard) to satisfy Agent 1 when its compatible users are rare, the system is able to achieve better fairness for both agents. 

\begin{figure}
    \centering
    \includegraphics[width=0.50\textwidth]{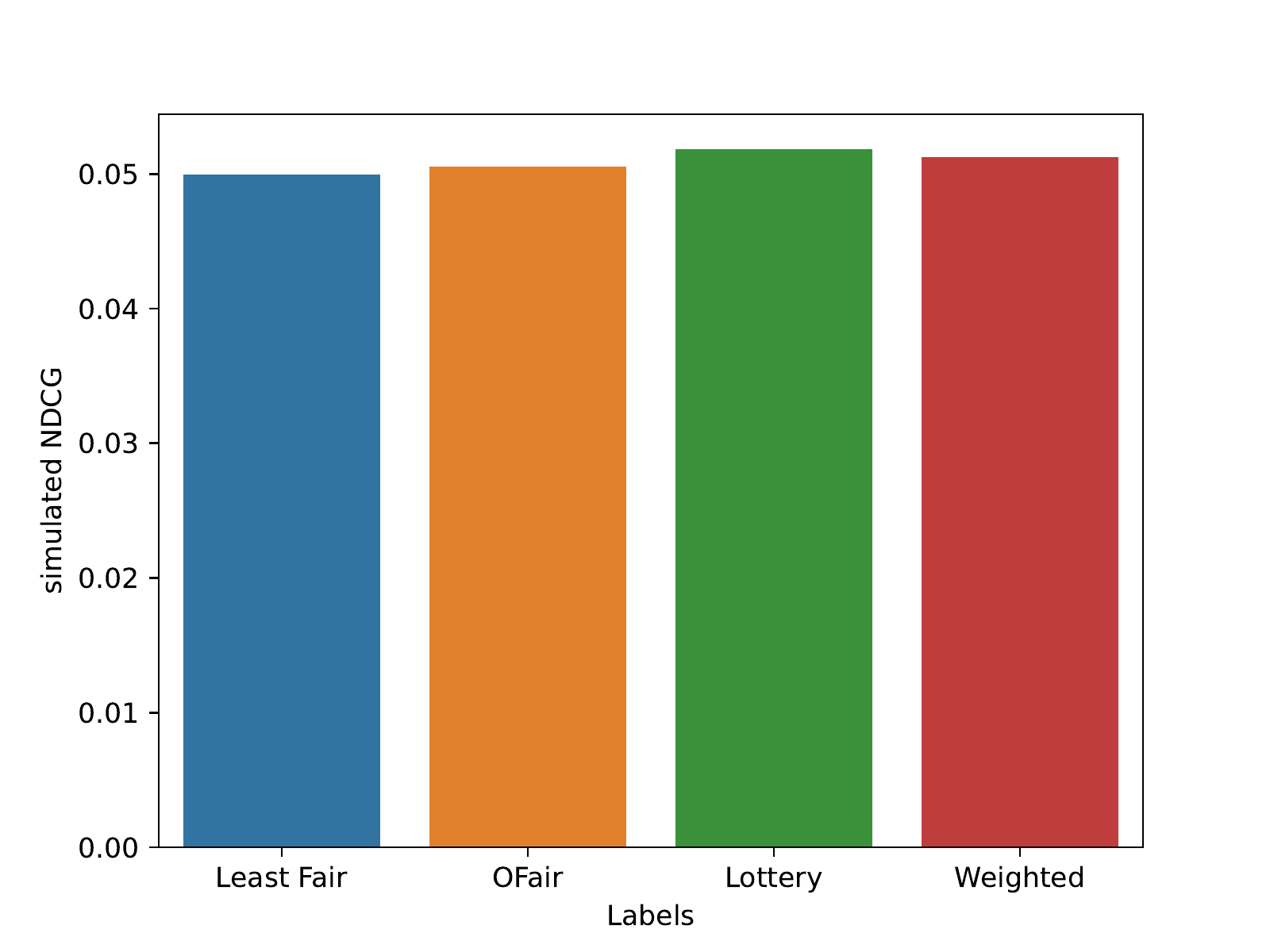}
    \caption{Ranking accuracy loss on the Synthetic data.}
    \label{fig:synthetic-ndcg}
\end{figure}

The Least Fair mechanism seems even better still for both agents. However, there is more to the story. By ignoring compatibility, this mechanism cannot achieve ranking accuracy as high as the others. This point is supported by Figure~\ref{fig:synthetic-ndcg}, which shows the loss of accuracy incurred by each algorithm. These are small differences, but the Least Fair mechanism is worst on this metric.  

\section{Conclusion and Future Work}
We have introduced the SCRUF-D architecture for integrating multiple fairness concerns into recommendation generation by leveraging social choice. The design is general and allows for many different types of fairness concerns: involving multiple fairness logics and encompassing both provider and consumer aspects of the recommendation platform. The architecture is also general in that it makes few assumptions about the nature of the allocation and choice mechanisms by which fairness is maintained, allowing for a large design space incorporating many types of functions.

Our experiments represent a first step at exploring the interactions of different allocation and choice mechanisms at the heart of the SCRUF model. We have found that the interaction between mechanisms is quite data and application specific although some general patterns emerge. A simple Rescoring technique is often just as good or better than more complex choice mechanisms. The Least Fair allocation mechanism, which ignores user preferences in allocating agents, is in some cases quite competitive with more sensitive allocations but some times incurs a substantial accuracy loss as we saw in the Microlending dataset.

Our experiments with synthetic data show that the SCRUF-D architecture is capable of balancing among multiple fairness concerns dynamically and in the end, much better fairness results can be achieved by dynamic mechanisms able to respond to the current fairness needs of each agent, as opposed to the static approaches seen in OFair and Multi-FR.

Future work will proceed in multiple research arcs. One arc of future work is to apply the architecture in more realistic settings, particularly with Kiva. We are working with Kiva stakeholders and beginning the process of formalizing fairness concerns as documented in \cite{smith2023many}. 

We have made the mechanisms and the agents fairly simple by design. Further experimentation will show how effective this structure is for maintaining fairness over time and allowing a wide variety of fairness concerns to be expressed. However, there are some areas of exploration that we can anticipate.

Our experiments raise the question of the impact of the base recommender and data characteristics on potential outcomes. We will explore additional choices for recommendation algorithms and datasets to explore and confirm the findings here.

For reasons of space, some key variations on the experiments shown here were not explored. All of the experiments contain only two fairness agents, although this is not a limitation of the architecture. It will be important to see how the results found here extend to larger cohorts of agents. Similarly, we have limited our agent definitions so that all agents have the a similar fairness definition (targeted exposure). We will explore more diverse and heterogenous fairness definitions across agents in future work. We also note that our synthetic data experiments were limited only to examining user arrival sequencing but with a flexible data generation scheme, there are many additional variables to explore including studying how well niche users are served.

We note that in a recommendation context the decisions of the recommender system only influence the exposure of protected items. There is no guarantee that a given user will show any interest in an item just because it is presented. In some settings and for some fairness concerns, exposure might be enough. But in cases where utility derives from usage rather than exposure, there would be some value in having the system learn about the relationship between exposure and utility. This setting has the attributes of a multi-objective bandit learning problem \cite{mehrotra2020bandit}, where the fairness concerns represent different classes of rewards and the allocation of agents represents different choices. 

Even when we consider exposure as our main outcome of interest, it is still the case that the allocation of different agents may result in differential improvements in fairness, the efficacy problem noted above. Perhaps the items associated with one agent are more common in recommendation lists and can be easily promoted through re-ranking while other agents' items are not. The weight associated with the allocation of agents may need to be adjusted to reflect the expected utility of allocation, and this expected utility would need to be learned.

The current architecture does not make any assumptions about the distribution of user characteristics and this can reduce its effectiveness. Suppose fairness concern $f_i$ is ``difficult'' to achieve in that users with an interest in related items appear rarely. In that case, we should probably allocate $f_i$ whenever a compatible user arrives, regardless of the state of the fairness metrics. This example suggests that the allocation mechanism could be adapted to look forward (to the distribution of future opportunities) as well as backwards (over fairness results achieved). This would require a model of opportunities similar to \cite{perlich2012bid}, and others studied in computational advertising settings.

The current architecture envisions fairness primarily in the context of group fairness expressed over recommendation outcomes. We believe that the architecture will support other types of fairness with additional enhancements. For example, a representational fairness concern would be incompatible with the assumption that fairness can be aggregated over multiple recommendation lists. Consider the examples in Noble's \textit{Algorithms of Oppression}: it would not be acceptable for a recommender system to deliver results that reinforced racist or sexist stereotypes at times, even if those results were balanced out at other times in some overall average. Representational fairness imposes a stricter constraint than those considered here, effectively requiring that the associated concern be allocated for every recommendation opportunity. 

As noted above, the model expressed here assumes that fairness agents have preferences only over items. But it is also possible to represent agents as having preferences over recommendation lists. This would allow agents to express preferences for combinations of items: for example, a preference that there be at least two Agriculture loans in the top 5 items of the list. This kind of preference cannot be expressed simply in terms of scores associated with items. Agents would naturally have to become more complex in their ability to reason about and generate such preferences, and the choice mechanism would become more like a combinatorial optimization problem. It is possible that we can characterize useful subclasses of the permutation space and avoid the full complexity of arbitrary preferences over subsets. 

Another interesting direction for research is more theoretical in nature. Much of the research in social choice focuses on providing guaranteed normative properties of various mechanisms. However, the models used in traditional social choice theory do not take into consideration the dynamics of recommender systems as most mechanisms are designed to work in one-off scenarios without dynamic aspects. One direction would be to formulate the allocation phase of the architecture as an online matching problem, where fairness agents represent one side of the matching and users arrive online on the other side, revealing their compatibility metric. Similar to work in online ad allocation, each fairness agent might have some budget or capacity that limits the number of users they are matched with, in order to balance between various fairness concerns. It will be important to understand the properties of existing social choice mechanisms for allocation and choice when deployed in these dynamic contexts and to develop new methods with good properties.
\color{black}
\begin{acks}
Authors Burke, Voida and Aird were supported by the National Science Foundation under grant awards IIS-1911025 and IIS-2107577. Nicholas Mattei was supported by NSF Grant IIS-2107505. Many thanks to Pradeep Ragothaman and our other collaborators at Kiva for sharing their data and many insights into their important work supporting international development through microlending. Thanks also to Lalita Suwattee for her assistance in conducting experiments.
\end{acks}

\bibliographystyle{ACM-Reference-Format}
\bibliography{scruff}


\begin{thebibliography}{63}


\ifx \showCODEN    \undefined \def \showCODEN     #1{\unskip}     \fi
\ifx \showDOI      \undefined \def \showDOI       #1{#1}\fi
\ifx \showISBNx    \undefined \def \showISBNx     #1{\unskip}     \fi
\ifx \showISBNxiii \undefined \def \showISBNxiii  #1{\unskip}     \fi
\ifx \showISSN     \undefined \def \showISSN      #1{\unskip}     \fi
\ifx \showLCCN     \undefined \def \showLCCN      #1{\unskip}     \fi
\ifx \shownote     \undefined \def \shownote      #1{#1}          \fi
\ifx \showarticletitle \undefined \def \showarticletitle #1{#1}   \fi
\ifx \showURL      \undefined \def \showURL       {\relax}        \fi
\providecommand\bibfield[2]{#2}
\providecommand\bibinfo[2]{#2}
\providecommand\natexlab[1]{#1}
\providecommand\showeprint[2][]{arXiv:#2}

\bibitem[\protect\citeauthoryear{Akpinar, DiCiccio, Nandy, and Basu}{Akpinar
  et~al\mbox{.}}{2022}]%
        {akpinar2022long}
\bibfield{author}{\bibinfo{person}{Nil-Jana Akpinar}, \bibinfo{person}{Cyrus
  DiCiccio}, \bibinfo{person}{Preetam Nandy}, {and} \bibinfo{person}{Kinjal
  Basu}.} \bibinfo{year}{2022}\natexlab{}.
\newblock \showarticletitle{Long-term Dynamics of Fairness Intervention in
  Connection Recommender Systems}. In \bibinfo{booktitle}{\emph{Proceedings of
  the 2022 AAAI/ACM Conference on AI, Ethics, and Society}}.
  \bibinfo{pages}{22--35}.
\newblock


\bibitem[\protect\citeauthoryear{Aleksandrov, Aziz, Gaspers, and
  Walsh}{Aleksandrov et~al\mbox{.}}{2015}]%
        {AAGW15a}
\bibfield{author}{\bibinfo{person}{M. Aleksandrov}, \bibinfo{person}{H. Aziz},
  \bibinfo{person}{S. Gaspers}, {and} \bibinfo{person}{T. Walsh}.}
  \bibinfo{year}{2015}\natexlab{}.
\newblock \showarticletitle{Online Fair Division: {A}nalysing a Food Bank
  problem}. In \bibinfo{booktitle}{\emph{Proc. 24th International Joint
  Conference on AI (IJCAI)}}. \bibinfo{publisher}{IJCAI},
  \bibinfo{pages}{2540--2546}.
\newblock


\bibitem[\protect\citeauthoryear{Amanatidis, Aziz, Birmpas, Filos-Ratsikas, Li,
  Moulin, Voudouris, and Wu}{Amanatidis et~al\mbox{.}}{2023}]%
        {amanatidis2023fair}
\bibfield{author}{\bibinfo{person}{Georgios Amanatidis}, \bibinfo{person}{Haris
  Aziz}, \bibinfo{person}{Georgios Birmpas}, \bibinfo{person}{Aris
  Filos-Ratsikas}, \bibinfo{person}{Bo Li}, \bibinfo{person}{Herv{\'e} Moulin},
  \bibinfo{person}{Alexandros~A Voudouris}, {and} \bibinfo{person}{Xiaowei
  Wu}.} \bibinfo{year}{2023}\natexlab{}.
\newblock \showarticletitle{Fair Division of Indivisible Goods: Recent Progress
  and Open Questions}.
\newblock \bibinfo{journal}{\emph{Artificial Intelligence}}
  (\bibinfo{year}{2023}), \bibinfo{pages}{103965}.
\newblock


\bibitem[\protect\citeauthoryear{Awasthi and Sandholm}{Awasthi and
  Sandholm}{2009}]%
        {AwSa09a}
\bibfield{author}{\bibinfo{person}{P. Awasthi} {and} \bibinfo{person}{T.
  Sandholm}.} \bibinfo{year}{2009}\natexlab{}.
\newblock \showarticletitle{Online Stochastic Optimization in the Large:
  Application to Kidney Exchange.}. In \bibinfo{booktitle}{\emph{Proc. 21st
  International Joint Conference on AI (IJCAI)}}. \bibinfo{publisher}{IJCAI},
  \bibinfo{pages}{405--411}.
\newblock


\bibitem[\protect\citeauthoryear{Barocas and Selbst}{Barocas and
  Selbst}{2016}]%
        {Barocas2016-wi}
\bibfield{author}{\bibinfo{person}{Solon Barocas} {and}
  \bibinfo{person}{Andrew~D Selbst}.} \bibinfo{year}{2016}\natexlab{}.
\newblock \showarticletitle{Big Data's Disparate Impact}.
\newblock \bibinfo{journal}{\emph{California law review}}
  \bibinfo{volume}{104}, \bibinfo{number}{3} (\bibinfo{year}{2016}),
  \bibinfo{pages}{671}.
\newblock
\showISSN{0008-1221}
\urldef\tempurl%
\url{https://doi.org/10.15779/Z38BG31}
\showDOI{\tempurl}


\bibitem[\protect\citeauthoryear{Bogomolnaia and Moulin}{Bogomolnaia and
  Moulin}{2001}]%
        {bogomolnaia2001new}
\bibfield{author}{\bibinfo{person}{Anna Bogomolnaia} {and}
  \bibinfo{person}{Herv{\'e} Moulin}.} \bibinfo{year}{2001}\natexlab{}.
\newblock \showarticletitle{A new solution to the random assignment problem}.
\newblock \bibinfo{journal}{\emph{Journal of Economic theory}}
  \bibinfo{volume}{100}, \bibinfo{number}{2} (\bibinfo{year}{2001}),
  \bibinfo{pages}{295--328}.
\newblock


\bibitem[\protect\citeauthoryear{Brandt, Conitzer, Endriss, Lang, and
  Procaccia}{Brandt et~al\mbox{.}}{2016}]%
        {BCELP16a}
\bibfield{editor}{\bibinfo{person}{F. Brandt}, \bibinfo{person}{V. Conitzer},
  \bibinfo{person}{U. Endriss}, \bibinfo{person}{J. Lang}, {and}
  \bibinfo{person}{A.~D. Procaccia}} (Eds.). \bibinfo{year}{2016}\natexlab{}.
\newblock \bibinfo{booktitle}{\emph{Handbook of Computational Social Choice}}.
\newblock \bibinfo{publisher}{Cambridge University Press}.
\newblock


\bibitem[\protect\citeauthoryear{Budish, Che, Kojima, and Milgrom}{Budish
  et~al\mbox{.}}{2013}]%
        {budish2013designing}
\bibfield{author}{\bibinfo{person}{Eric Budish}, \bibinfo{person}{Yeon-Koo
  Che}, \bibinfo{person}{Fuhito Kojima}, {and} \bibinfo{person}{Paul Milgrom}.}
  \bibinfo{year}{2013}\natexlab{}.
\newblock \showarticletitle{Designing random allocation mechanisms: Theory and
  applications}.
\newblock \bibinfo{journal}{\emph{American economic review}}
  \bibinfo{volume}{103}, \bibinfo{number}{2} (\bibinfo{year}{2013}),
  \bibinfo{pages}{585--623}.
\newblock


\bibitem[\protect\citeauthoryear{Buet-Golfouse and Utyagulov}{Buet-Golfouse and
  Utyagulov}{2022}]%
        {buet2022towards}
\bibfield{author}{\bibinfo{person}{Francois Buet-Golfouse} {and}
  \bibinfo{person}{Islam Utyagulov}.} \bibinfo{year}{2022}\natexlab{}.
\newblock \showarticletitle{Towards fair multi-stakeholder recommender
  systems}. In \bibinfo{booktitle}{\emph{Adjunct Proceedings of the 30th ACM
  Conference on User Modeling, Adaptation and Personalization}}.
  \bibinfo{pages}{255--265}.
\newblock


\bibitem[\protect\citeauthoryear{Burke}{Burke}{2017}]%
        {burke_multisided_2017}
\bibfield{author}{\bibinfo{person}{Robin Burke}.}
  \bibinfo{year}{2017}\natexlab{}.
\newblock \showarticletitle{Multisided {Fairness} for {Recommendation}}. In
  \bibinfo{booktitle}{\emph{Workshop on {Fairness}, {Accountability} and
  {Transparency} in {Machine} {Learning} ({FATML})}}.
  \bibinfo{address}{Halifax, Nova Scotia}, \bibinfo{numpages}{5}~pages.
\newblock
\urldef\tempurl%
\url{https://arxiv.org/abs/1707.00093}
\showURL{%
\tempurl}


\bibitem[\protect\citeauthoryear{Burke, Mattei, Grozin, Voida, and
  Sonboli}{Burke et~al\mbox{.}}{2022a}]%
        {burke2022multi}
\bibfield{author}{\bibinfo{person}{Robin Burke}, \bibinfo{person}{Nicholas
  Mattei}, \bibinfo{person}{Vladislav Grozin}, \bibinfo{person}{Amy Voida},
  {and} \bibinfo{person}{Nasim Sonboli}.} \bibinfo{year}{2022}\natexlab{a}.
\newblock \showarticletitle{Multi-agent Social Choice for Dynamic
  Fairness-aware Recommendation}. In \bibinfo{booktitle}{\emph{Adjunct
  Proceedings of the 30th ACM Conference on User Modeling, Adaptation and
  Personalization}}. \bibinfo{pages}{234--244}.
\newblock


\bibitem[\protect\citeauthoryear{Burke, Ragothaman, Mattei, Kimmig, Voida,
  Sonboli, Kathait, and Fabros}{Burke et~al\mbox{.}}{2022b}]%
        {burke2022performance}
\bibfield{author}{\bibinfo{person}{Robin Burke}, \bibinfo{person}{Pradeep
  Ragothaman}, \bibinfo{person}{Nicholas Mattei}, \bibinfo{person}{Brian
  Kimmig}, \bibinfo{person}{Amy Voida}, \bibinfo{person}{Nasim Sonboli},
  \bibinfo{person}{Anushka Kathait}, {and} \bibinfo{person}{Melissa Fabros}.}
  \bibinfo{year}{2022}\natexlab{b}.
\newblock \showarticletitle{A performance-preserving fairness intervention for
  adaptive microfinance recommendation}. In
  \bibinfo{booktitle}{\emph{Proceedings of the KDD Workshop on Online and
  Adapting Recommender Systems (OARS)}}.
\newblock


\bibitem[\protect\citeauthoryear{Burke, Sonboli, and Ordonez-Gauger}{Burke
  et~al\mbox{.}}{2018}]%
        {burke2018balanced}
\bibfield{author}{\bibinfo{person}{Robin Burke}, \bibinfo{person}{Nasim
  Sonboli}, {and} \bibinfo{person}{Aldo Ordonez-Gauger}.}
  \bibinfo{year}{2018}\natexlab{}.
\newblock \showarticletitle{Balanced Neighborhoods for Multi-sided Fairness in
  Recommendation}. In \bibinfo{booktitle}{\emph{Proceedings of the 1st
  Conference on Fairness, Accountability and Transparency}}
  \emph{(\bibinfo{series}{Proceedings of Machine Learning Research},
  Vol.~\bibinfo{volume}{81})}, \bibfield{editor}{\bibinfo{person}{Sorelle~A.
  Friedler} {and} \bibinfo{person}{Christo Wilson}} (Eds.).
  \bibinfo{publisher}{PMLR}, \bibinfo{address}{New York, NY, USA},
  \bibinfo{pages}{202--214}.
\newblock


\bibitem[\protect\citeauthoryear{Burke, Voida, Mattei, and Sonboli}{Burke
  et~al\mbox{.}}{2020}]%
        {Burke:AlgoFairness}
\bibfield{author}{\bibinfo{person}{Robin Burke}, \bibinfo{person}{Amy Voida},
  \bibinfo{person}{Nicholas Mattei}, {and} \bibinfo{person}{Nasim Sonboli}.}
  \bibinfo{year}{2020}\natexlab{}.
\newblock \showarticletitle{Algorithmic Fairness, Institutional Logics, and
  Social Choice}. In \bibinfo{booktitle}{\emph{Harvard CRCS Workshop on AI for
  Social Good at 29th International Joint Conference on Artificial Intelligence
  {(IJCAI 2020)}}}. \bibinfo{numpages}{5}~pages.
\newblock


\bibitem[\protect\citeauthoryear{Carbonell and Goldstein}{Carbonell and
  Goldstein}{1998}]%
        {carbonell1998use}
\bibfield{author}{\bibinfo{person}{Jaime Carbonell} {and} \bibinfo{person}{Jade
  Goldstein}.} \bibinfo{year}{1998}\natexlab{}.
\newblock \showarticletitle{The use of MMR, diversity-based reranking for
  reordering documents and producing summaries}. In
  \bibinfo{booktitle}{\emph{Proceedings of the 21st annual international ACM
  SIGIR conference on Research and development in information retrieval}}.
  \bibinfo{pages}{335--336}.
\newblock


\bibitem[\protect\citeauthoryear{Cesa-Bianchi, Freund, Haussler, Helmbold,
  Schapire, and Warmuth}{Cesa-Bianchi et~al\mbox{.}}{1997}]%
        {cesa1997use}
\bibfield{author}{\bibinfo{person}{Nicolo Cesa-Bianchi}, \bibinfo{person}{Yoav
  Freund}, \bibinfo{person}{David Haussler}, \bibinfo{person}{David~P
  Helmbold}, \bibinfo{person}{Robert~E Schapire}, {and}
  \bibinfo{person}{Manfred~K Warmuth}.} \bibinfo{year}{1997}\natexlab{}.
\newblock \showarticletitle{How to use expert advice}.
\newblock \bibinfo{journal}{\emph{Journal of the ACM (JACM)}}
  \bibinfo{volume}{44}, \bibinfo{number}{3} (\bibinfo{year}{1997}),
  \bibinfo{pages}{427--485}.
\newblock


\bibitem[\protect\citeauthoryear{Cohler, Lai, Parkes, and Procaccia}{Cohler
  et~al\mbox{.}}{2011}]%
        {cohler2011optimal}
\bibfield{author}{\bibinfo{person}{Yuga~J Cohler}, \bibinfo{person}{John~K
  Lai}, \bibinfo{person}{David~C Parkes}, {and} \bibinfo{person}{Ariel~D
  Procaccia}.} \bibinfo{year}{2011}\natexlab{}.
\newblock \showarticletitle{Optimal envy-free cake cutting}. In
  \bibinfo{booktitle}{\emph{Twenty-Fifth AAAI Conference on Artificial
  Intelligence}}.
\newblock


\bibitem[\protect\citeauthoryear{Dickerson, Sankararaman, Srinivasan, and
  Xu}{Dickerson et~al\mbox{.}}{2018}]%
        {dickerson2018allocation}
\bibfield{author}{\bibinfo{person}{John~P Dickerson},
  \bibinfo{person}{Karthik~A Sankararaman}, \bibinfo{person}{Aravind
  Srinivasan}, {and} \bibinfo{person}{Pan Xu}.}
  \bibinfo{year}{2018}\natexlab{}.
\newblock \showarticletitle{Allocation Problems in Ride Sharing Platforms:
  Online Matching with Offline Reusable Resources}. In
  \bibinfo{booktitle}{\emph{Proc. Thirty-Second AAAI Conference on Artificial
  Intelligence (AAAI)}}. \bibinfo{publisher}{AAAI},
  \bibinfo{pages}{1007--1014}.
\newblock


\bibitem[\protect\citeauthoryear{Edelman, Ostrovsky, and Schwarz}{Edelman
  et~al\mbox{.}}{2007}]%
        {edelman2007internet}
\bibfield{author}{\bibinfo{person}{Benjamin Edelman}, \bibinfo{person}{Michael
  Ostrovsky}, {and} \bibinfo{person}{Michael Schwarz}.}
  \bibinfo{year}{2007}\natexlab{}.
\newblock \showarticletitle{Internet advertising and the generalized
  second-price auction: Selling billions of dollars worth of keywords}.
\newblock \bibinfo{journal}{\emph{The American economic review}}
  \bibinfo{volume}{97}, \bibinfo{number}{1} (\bibinfo{year}{2007}),
  \bibinfo{pages}{242--259}.
\newblock


\bibitem[\protect\citeauthoryear{Ekstrand, Das, Burke, and Diaz}{Ekstrand
  et~al\mbox{.}}{2022}]%
        {ekstrand2022fairness}
\bibfield{author}{\bibinfo{person}{Michael~D. Ekstrand},
  \bibinfo{person}{Anubrata Das}, \bibinfo{person}{Robin Burke}, {and}
  \bibinfo{person}{Fernando Diaz}.} \bibinfo{year}{2022}\natexlab{}.
\newblock \bibinfo{title}{Fairness in Information Access Systems}.
\newblock
\newblock
\showeprint[arxiv]{2105.05779}~[cs.IR]


\bibitem[\protect\citeauthoryear{Farastu, Mattei, and Burke}{Farastu
  et~al\mbox{.}}{2022}]%
        {farastu2022pays}
\bibfield{author}{\bibinfo{person}{Paresha Farastu}, \bibinfo{person}{Nicholas
  Mattei}, {and} \bibinfo{person}{Robin Burke}.}
  \bibinfo{year}{2022}\natexlab{}.
\newblock \showarticletitle{Who Pays? Personalization, Bossiness and the Cost
  of Fairness}.
\newblock \bibinfo{journal}{\emph{arXiv preprint arXiv:2209.04043}}
  (\bibinfo{year}{2022}).
\newblock


\bibitem[\protect\citeauthoryear{Ferraro, Serra, and Bauer}{Ferraro
  et~al\mbox{.}}{2021}]%
        {ferraro2021break}
\bibfield{author}{\bibinfo{person}{Andres Ferraro}, \bibinfo{person}{Xavier
  Serra}, {and} \bibinfo{person}{Christine Bauer}.}
  \bibinfo{year}{2021}\natexlab{}.
\newblock \showarticletitle{Break the loop: Gender imbalance in music
  recommenders}. In \bibinfo{booktitle}{\emph{Proceedings of the 2021
  Conference on Human Information Interaction and Retrieval}}.
  \bibinfo{pages}{249--254}.
\newblock


\bibitem[\protect\citeauthoryear{Freeman, Zahedi, and Conitzer}{Freeman
  et~al\mbox{.}}{2017}]%
        {freeman2017fair}
\bibfield{author}{\bibinfo{person}{Rupert Freeman},
  \bibinfo{person}{Seyed~Majid Zahedi}, {and} \bibinfo{person}{Vincent
  Conitzer}.} \bibinfo{year}{2017}\natexlab{}.
\newblock \showarticletitle{Fair social choice in dynamic settings}. In
  \bibinfo{booktitle}{\emph{Proceedings of the 26th International Joint
  Conference on Artificial Intelligence (IJCAI)}}.
  \bibinfo{publisher}{International Joint Conferences on Artificial
  Intelligence}, \bibinfo{address}{Marina del Rey, CA},
  \bibinfo{pages}{4580--4587}.
\newblock


\bibitem[\protect\citeauthoryear{Friedler, Scheidegger, and
  Venkatasubramanian}{Friedler et~al\mbox{.}}{2021}]%
        {friedler2021possibility}
\bibfield{author}{\bibinfo{person}{Sorelle~A Friedler}, \bibinfo{person}{Carlos
  Scheidegger}, {and} \bibinfo{person}{Suresh Venkatasubramanian}.}
  \bibinfo{year}{2021}\natexlab{}.
\newblock \showarticletitle{The (Im)possibility of fairness: different value
  systems require different mechanisms for fair decision making}.
\newblock \bibinfo{journal}{\emph{Commun. ACM}} \bibinfo{volume}{64},
  \bibinfo{number}{4} (\bibinfo{date}{April} \bibinfo{year}{2021}),
  \bibinfo{pages}{136--143}.
\newblock
\showISSN{0001-0782, 1557-7317}
\urldef\tempurl%
\url{https://doi.org/10.1145/3433949}
\showDOI{\tempurl}


\bibitem[\protect\citeauthoryear{Ge, Liu, Gao, Xian, Li, Zhao, Pei, Sun, Ge,
  Ou, et~al\mbox{.}}{Ge et~al\mbox{.}}{2021}]%
        {ge2021towards}
\bibfield{author}{\bibinfo{person}{Yingqiang Ge}, \bibinfo{person}{Shuchang
  Liu}, \bibinfo{person}{Ruoyuan Gao}, \bibinfo{person}{Yikun Xian},
  \bibinfo{person}{Yunqi Li}, \bibinfo{person}{Xiangyu Zhao},
  \bibinfo{person}{Changhua Pei}, \bibinfo{person}{Fei Sun},
  \bibinfo{person}{Junfeng Ge}, \bibinfo{person}{Wenwu Ou}, {et~al\mbox{.}}}
  \bibinfo{year}{2021}\natexlab{}.
\newblock \showarticletitle{Towards long-term fairness in recommendation}. In
  \bibinfo{booktitle}{\emph{Proceedings of the 14th ACM International
  Conference on Web Search and Data Mining}}. \bibinfo{publisher}{ACM},
  \bibinfo{address}{New York}, \bibinfo{pages}{445--453}.
\newblock


\bibitem[\protect\citeauthoryear{Harper and Konstan}{Harper and
  Konstan}{2015}]%
        {movielens}
\bibfield{author}{\bibinfo{person}{F~Maxwell Harper} {and}
  \bibinfo{person}{Joseph~A Konstan}.} \bibinfo{year}{2015}\natexlab{}.
\newblock \showarticletitle{The MovieLens Datasets: History and Context}.
\newblock \bibinfo{journal}{\emph{ACM Transactions on Interactive Intelligent
  Systems (TiiS)}} \bibinfo{volume}{5}, \bibinfo{number}{4}
  (\bibinfo{year}{2015}), \bibinfo{pages}{19}.
\newblock


\bibitem[\protect\citeauthoryear{Hutchinson and Mitchell}{Hutchinson and
  Mitchell}{2019}]%
        {Hutchinson_2019}
\bibfield{author}{\bibinfo{person}{Ben Hutchinson} {and}
  \bibinfo{person}{Margaret Mitchell}.} \bibinfo{year}{2019}\natexlab{}.
\newblock \showarticletitle{50 Years of Test (Un)fairness}.
\newblock \bibinfo{journal}{\emph{Proceedings of the Conference on Fairness,
  Accountability, and Transparency - FAT* '19}} (\bibinfo{year}{2019}).
\newblock


\bibitem[\protect\citeauthoryear{Jannach, Lerche, Kamehkhosh, and
  Jugovac}{Jannach et~al\mbox{.}}{2015}]%
        {jannach2015recommenders}
\bibfield{author}{\bibinfo{person}{Dietmar Jannach}, \bibinfo{person}{Lukas
  Lerche}, \bibinfo{person}{Iman Kamehkhosh}, {and} \bibinfo{person}{Michael
  Jugovac}.} \bibinfo{year}{2015}\natexlab{}.
\newblock \showarticletitle{What recommenders recommend: an analysis of
  recommendation biases and possible countermeasures}.
\newblock \bibinfo{journal}{\emph{User Modeling and User-Adapted Interaction}}
  \bibinfo{volume}{25} (\bibinfo{year}{2015}), \bibinfo{pages}{427--491}.
\newblock


\bibitem[\protect\citeauthoryear{Kaya, Bridge, and Tintarev}{Kaya
  et~al\mbox{.}}{2020}]%
        {kaya2020ensuring}
\bibfield{author}{\bibinfo{person}{Mesut Kaya}, \bibinfo{person}{Derek Bridge},
  {and} \bibinfo{person}{Nava Tintarev}.} \bibinfo{year}{2020}\natexlab{}.
\newblock \showarticletitle{Ensuring fairness in group recommendations by
  rank-sensitive balancing of relevance}. In
  \bibinfo{booktitle}{\emph{Proceedings of the 14th ACM Conference on
  Recommender Systems}}. \bibinfo{pages}{101--110}.
\newblock


\bibitem[\protect\citeauthoryear{Kearns, Neel, Roth, and Wu}{Kearns
  et~al\mbox{.}}{2018}]%
        {kearns2018preventing}
\bibfield{author}{\bibinfo{person}{Michael Kearns}, \bibinfo{person}{Seth
  Neel}, \bibinfo{person}{Aaron Roth}, {and} \bibinfo{person}{Zhiwei~Steven
  Wu}.} \bibinfo{year}{2018}\natexlab{}.
\newblock \bibinfo{title}{Preventing Fairness Gerrymandering: Auditing and
  Learning for Subgroup Fairness}.
\newblock
\newblock
\showeprint[arxiv]{1711.05144}~[cs.LG]


\bibitem[\protect\citeauthoryear{Lee, Kusbit, Kahng, Kim, Yuan, Chan, See,
  Noothigattu, Lee, and Psomas}{Lee et~al\mbox{.}}{2019}]%
        {lee2019webuildai}
\bibfield{author}{\bibinfo{person}{Min~Kyung Lee}, \bibinfo{person}{Daniel
  Kusbit}, \bibinfo{person}{Anson Kahng}, \bibinfo{person}{Ji~Tae Kim},
  \bibinfo{person}{Xinran Yuan}, \bibinfo{person}{Allissa Chan},
  \bibinfo{person}{Daniel See}, \bibinfo{person}{Ritesh Noothigattu},
  \bibinfo{person}{Siheon Lee}, {and} \bibinfo{person}{Alexandros Psomas}.}
  \bibinfo{year}{2019}\natexlab{}.
\newblock \showarticletitle{WeBuildAI: Participatory framework for algorithmic
  governance}.
\newblock \bibinfo{journal}{\emph{Proceedings of the ACM on Human-Computer
  Interaction}} \bibinfo{volume}{3}, \bibinfo{number}{CSCW}
  (\bibinfo{year}{2019}), \bibinfo{pages}{1--35}.
\newblock


\bibitem[\protect\citeauthoryear{Li, Hsu, and Zhang}{Li et~al\mbox{.}}{2022}]%
        {li2022fairsr}
\bibfield{author}{\bibinfo{person}{Cheng-Te Li}, \bibinfo{person}{Cheng Hsu},
  {and} \bibinfo{person}{Yang Zhang}.} \bibinfo{year}{2022}\natexlab{}.
\newblock \showarticletitle{Fair{SR}: Fairness-aware sequential recommendation
  through multi-task learning with preference graph embeddings}.
\newblock \bibinfo{journal}{\emph{ACM Transactions on Intelligent Systems and
  Technology (TIST)}} \bibinfo{volume}{13}, \bibinfo{number}{1}
  (\bibinfo{year}{2022}), \bibinfo{pages}{1--21}.
\newblock


\bibitem[\protect\citeauthoryear{Liu and Burke}{Liu and Burke}{2018}]%
        {liu2018personalizing}
\bibfield{author}{\bibinfo{person}{Weiwen Liu} {and} \bibinfo{person}{Robin
  Burke}.} \bibinfo{year}{2018}\natexlab{}.
\newblock \showarticletitle{Personalizing Fairness-aware Re-ranking}.
\newblock \bibinfo{journal}{\emph{arXiv preprint arXiv:1809.02921}}
  (\bibinfo{year}{2018}).
\newblock
\newblock
\shownote{Presented at the 2nd FATRec Workshop held at RecSys 2018, Vancouver,
  CA.}.


\bibitem[\protect\citeauthoryear{Mattei, Saffidine, and Walsh}{Mattei
  et~al\mbox{.}}{2018}]%
        {MaSaWa18Com}
\bibfield{author}{\bibinfo{person}{N. Mattei}, \bibinfo{person}{A. Saffidine},
  {and} \bibinfo{person}{T. Walsh}.} \bibinfo{year}{2018}\natexlab{}.
\newblock \showarticletitle{An Axiomatic and Empirical Analysis of Mechanisms
  for Online Organ Matching}. In \bibinfo{booktitle}{\emph{Proceedings of the
  7th International Workshop on Computational Social Choice (COMSOC)}}.
  \bibinfo{numpages}{24}~pages.
\newblock


\bibitem[\protect\citeauthoryear{Mehrotra, Xue, and Lalmas}{Mehrotra
  et~al\mbox{.}}{2020}]%
        {mehrotra2020bandit}
\bibfield{author}{\bibinfo{person}{Rishabh Mehrotra}, \bibinfo{person}{Niannan
  Xue}, {and} \bibinfo{person}{Mounia Lalmas}.}
  \bibinfo{year}{2020}\natexlab{}.
\newblock \showarticletitle{Bandit based Optimization of Multiple Objectives on
  a Music Streaming Platform}. In \bibinfo{booktitle}{\emph{Proceedings of the
  26th ACM SIGKDD International Conference on Knowledge Discovery \& Data
  Mining}}. \bibinfo{pages}{3224--3233}.
\newblock


\bibitem[\protect\citeauthoryear{Morik, Singh, Hong, and Joachims}{Morik
  et~al\mbox{.}}{2020}]%
        {morik2020controlling}
\bibfield{author}{\bibinfo{person}{Marco Morik}, \bibinfo{person}{Ashudeep
  Singh}, \bibinfo{person}{Jessica Hong}, {and} \bibinfo{person}{Thorsten
  Joachims}.} \bibinfo{year}{2020}\natexlab{}.
\newblock \showarticletitle{Controlling fairness and bias in dynamic
  learning-to-rank}. In \bibinfo{booktitle}{\emph{Proceedings of the 43rd
  International ACM SIGIR Conference on Research and Development in Information
  Retrieval}}. \bibinfo{pages}{429--438}.
\newblock


\bibitem[\protect\citeauthoryear{Moulin}{Moulin}{2004}]%
        {moulin2004fair}
\bibfield{author}{\bibinfo{person}{Herv{\'e} Moulin}.}
  \bibinfo{year}{2004}\natexlab{}.
\newblock \bibinfo{booktitle}{\emph{Fair division and collective welfare}}.
\newblock \bibinfo{publisher}{MIT press}.
\newblock


\bibitem[\protect\citeauthoryear{Mulligan, Kroll, Kohli, and Wong}{Mulligan
  et~al\mbox{.}}{2019}]%
        {mulligan2019thing}
\bibfield{author}{\bibinfo{person}{Deirdre~K Mulligan},
  \bibinfo{person}{Joshua~A Kroll}, \bibinfo{person}{Nitin Kohli}, {and}
  \bibinfo{person}{Richmond~Y Wong}.} \bibinfo{year}{2019}\natexlab{}.
\newblock \showarticletitle{This thing called fairness: disciplinary confusion
  realizing a value in technology}.
\newblock \bibinfo{journal}{\emph{Proceedings of the ACM on Human-Computer
  Interaction}} \bibinfo{volume}{3}, \bibinfo{number}{CSCW}
  (\bibinfo{year}{2019}), \bibinfo{pages}{1--36}.
\newblock


\bibitem[\protect\citeauthoryear{Noble}{Noble}{2018}]%
        {noble2018algorithms}
\bibfield{author}{\bibinfo{person}{Safiya~Umoja Noble}.}
  \bibinfo{year}{2018}\natexlab{}.
\newblock \bibinfo{booktitle}{\emph{Algorithms of Oppression: How search
  engines reinforce racism}}.
\newblock \bibinfo{publisher}{NYU Press}.
\newblock


\bibitem[\protect\citeauthoryear{O'Neil}{O'Neil}{2016}]%
        {o2016weapons}
\bibfield{author}{\bibinfo{person}{Cathy O'Neil}.}
  \bibinfo{year}{2016}\natexlab{}.
\newblock \bibinfo{booktitle}{\emph{Weapons of math destruction: How big data
  increases inequality and threatens democracy}}.
\newblock \bibinfo{publisher}{Broadway Books}.
\newblock


\bibitem[\protect\citeauthoryear{Pacuit}{Pacuit}{2019}]%
        {sep-voting-methods}
\bibfield{author}{\bibinfo{person}{Eric Pacuit}.}
  \bibinfo{year}{2019}\natexlab{}.
\newblock \showarticletitle{{Voting Methods}}.
\newblock In \bibinfo{booktitle}{\emph{The {Stanford} Encyclopedia of
  Philosophy} (\bibinfo{edition}{{F}all 2019} ed.)},
  \bibfield{editor}{\bibinfo{person}{Edward~N. Zalta}} (Ed.).
  \bibinfo{publisher}{Metaphysics Research Lab, Stanford University}.
\newblock


\bibitem[\protect\citeauthoryear{P{\'a}pai}{P{\'a}pai}{2000}]%
        {papai2000strategyproof}
\bibfield{author}{\bibinfo{person}{Szilvia P{\'a}pai}.}
  \bibinfo{year}{2000}\natexlab{}.
\newblock \showarticletitle{Strategyproof assignment by hierarchical exchange}.
\newblock \bibinfo{journal}{\emph{Econometrica}} \bibinfo{volume}{68},
  \bibinfo{number}{6} (\bibinfo{year}{2000}), \bibinfo{pages}{1403--1433}.
\newblock


\bibitem[\protect\citeauthoryear{Patro, Biswas, Ganguly, Gummadi, and
  Chakraborty}{Patro et~al\mbox{.}}{2020}]%
        {patro2020fairrec}
\bibfield{author}{\bibinfo{person}{Gourab~K Patro}, \bibinfo{person}{Arpita
  Biswas}, \bibinfo{person}{Niloy Ganguly}, \bibinfo{person}{Krishna~P
  Gummadi}, {and} \bibinfo{person}{Abhijnan Chakraborty}.}
  \bibinfo{year}{2020}\natexlab{}.
\newblock \showarticletitle{FairRec: Two-Sided Fairness for Personalized
  Recommendations in Two-Sided Platforms}. In
  \bibinfo{booktitle}{\emph{Proceedings of The Web Conference 2020}}.
  \bibinfo{pages}{1194--1204}.
\newblock


\bibitem[\protect\citeauthoryear{Patro, Porcaro, Mitchell, Zhang, Zehlike, and
  Garg}{Patro et~al\mbox{.}}{2022}]%
        {patro2022fair}
\bibfield{author}{\bibinfo{person}{Gourab~K Patro}, \bibinfo{person}{Lorenzo
  Porcaro}, \bibinfo{person}{Laura Mitchell}, \bibinfo{person}{Qiuyue Zhang},
  \bibinfo{person}{Meike Zehlike}, {and} \bibinfo{person}{Nikhil Garg}.}
  \bibinfo{year}{2022}\natexlab{}.
\newblock \showarticletitle{Fair ranking: a critical review, challenges, and
  future directions}. In \bibinfo{booktitle}{\emph{Proceedings of the 2022 ACM
  Conference on Fairness, Accountability, and Transparency}}.
  \bibinfo{pages}{1929--1942}.
\newblock


\bibitem[\protect\citeauthoryear{Perlich, Dalessandro, Hook, Stitelman, Raeder,
  and Provost}{Perlich et~al\mbox{.}}{2012}]%
        {perlich2012bid}
\bibfield{author}{\bibinfo{person}{Claudia Perlich}, \bibinfo{person}{Brian
  Dalessandro}, \bibinfo{person}{Rod Hook}, \bibinfo{person}{Ori Stitelman},
  \bibinfo{person}{Troy Raeder}, {and} \bibinfo{person}{Foster Provost}.}
  \bibinfo{year}{2012}\natexlab{}.
\newblock \showarticletitle{Bid optimizing and inventory scoring in targeted
  online advertising}. In \bibinfo{booktitle}{\emph{Proceedings of the 18th ACM
  SIGKDD international conference on Knowledge discovery and data mining}}.
  \bibinfo{pages}{804--812}.
\newblock


\bibitem[\protect\citeauthoryear{Smith, Beattie, and Cramer}{Smith
  et~al\mbox{.}}{2023a}]%
        {smith2023scoping}
\bibfield{author}{\bibinfo{person}{Jessie~J Smith}, \bibinfo{person}{Lex
  Beattie}, {and} \bibinfo{person}{Henriette Cramer}.}
  \bibinfo{year}{2023}\natexlab{a}.
\newblock \showarticletitle{Scoping Fairness Objectives and Identifying
  Fairness Metrics for Recommender Systems: The Practitioners’ Perspective}.
  In \bibinfo{booktitle}{\emph{Proceedings of the ACM Web Conference 2023}}.
  \bibinfo{pages}{3648--3659}.
\newblock


\bibitem[\protect\citeauthoryear{Smith, Buhayh, Kathait, Ragothaman, Mattei,
  Burke, and Voida}{Smith et~al\mbox{.}}{2023b}]%
        {smith2023many}
\bibfield{author}{\bibinfo{person}{Jessie~J Smith}, \bibinfo{person}{Anas
  Buhayh}, \bibinfo{person}{Anushka Kathait}, \bibinfo{person}{Pradeep
  Ragothaman}, \bibinfo{person}{Nicholas Mattei}, \bibinfo{person}{Robin
  Burke}, {and} \bibinfo{person}{Amy Voida}.} \bibinfo{year}{2023}\natexlab{b}.
\newblock \showarticletitle{The Many Faces of Fairness: Exploring the
  Institutional Logics of Multistakeholder Microlending Recommendation}. In
  \bibinfo{booktitle}{\emph{Proceedings of the 2023 ACM Conference on Fairness,
  Accountability, and Transparency}}. \bibinfo{pages}{1652--1663}.
\newblock


\bibitem[\protect\citeauthoryear{Sonboli, Aird, and Burke}{Sonboli
  et~al\mbox{.}}{2022}]%
        {sonboli2022micro}
\bibfield{author}{\bibinfo{person}{Nasim Sonboli}, \bibinfo{person}{Amanda
  Aird}, {and} \bibinfo{person}{Robin Burke}.} \bibinfo{year}{2022}\natexlab{}.
\newblock \bibinfo{booktitle}{\emph{Microlending 2017 Data Set}}.
\newblock
\urldef\tempurl%
\url{https://doi.org/10.25810/PGJK-RR19}
\showDOI{\tempurl}


\bibitem[\protect\citeauthoryear{Sonboli, Burke, Mattei, Eskandanian, and
  Gao}{Sonboli et~al\mbox{.}}{2020a}]%
        {sonboli2020and}
\bibfield{author}{\bibinfo{person}{Nasim Sonboli}, \bibinfo{person}{Robin
  Burke}, \bibinfo{person}{Nicholas Mattei}, \bibinfo{person}{Farzad
  Eskandanian}, {and} \bibinfo{person}{Tian Gao}.}
  \bibinfo{year}{2020}\natexlab{a}.
\newblock \bibinfo{title}{"And the Winner Is...": Dynamic Lotteries for
  Multi-group Fairness-Aware Recommendation}.
\newblock
\newblock
\showeprint[arxiv]{2009.02590}~[cs.IR]


\bibitem[\protect\citeauthoryear{Sonboli, Eskandanian, Burke, Liu, and
  Mobasher}{Sonboli et~al\mbox{.}}{2020b}]%
        {sonboli2020opportunistic}
\bibfield{author}{\bibinfo{person}{Nasim Sonboli}, \bibinfo{person}{Farzad
  Eskandanian}, \bibinfo{person}{Robin Burke}, \bibinfo{person}{Weiwen Liu},
  {and} \bibinfo{person}{Bamshad Mobasher}.} \bibinfo{year}{2020}\natexlab{b}.
\newblock \showarticletitle{Opportunistic Multi-Aspect Fairness through
  Personalized Re-Ranking}. In \bibinfo{booktitle}{\emph{Proceedings of the
  28th ACM Conference on User Modeling, Adaptation and Personalization}}
  (Genoa, Italy) \emph{(\bibinfo{series}{UMAP '20})}.
  \bibinfo{publisher}{Association for Computing Machinery},
  \bibinfo{address}{New York, NY, USA}, \bibinfo{pages}{239–247}.
\newblock
\showISBNx{9781450368612}
\urldef\tempurl%
\url{https://doi.org/10.1145/3340631.3394846}
\showDOI{\tempurl}


\bibitem[\protect\citeauthoryear{S{\"u}hr, Biega, Zehlike, Gummadi, and
  Chakraborty}{S{\"u}hr et~al\mbox{.}}{2019}]%
        {suhr2019two}
\bibfield{author}{\bibinfo{person}{Tom S{\"u}hr}, \bibinfo{person}{Asia~J
  Biega}, \bibinfo{person}{Meike Zehlike}, \bibinfo{person}{Krishna~P Gummadi},
  {and} \bibinfo{person}{Abhijnan Chakraborty}.}
  \bibinfo{year}{2019}\natexlab{}.
\newblock \showarticletitle{Two-sided fairness for repeated matchings in
  two-sided markets: A case study of a ride-hailing platform}. In
  \bibinfo{booktitle}{\emph{Proceedings of the 25th ACM SIGKDD International
  Conference on Knowledge Discovery \& Data Mining}}.
  \bibinfo{pages}{3082--3092}.
\newblock


\bibitem[\protect\citeauthoryear{Thomson}{Thomson}{2011}]%
        {Thomson:FairRules}
\bibfield{author}{\bibinfo{person}{William Thomson}.}
  \bibinfo{year}{2011}\natexlab{}.
\newblock \showarticletitle{Fair allocation rules}.
\newblock In \bibinfo{booktitle}{\emph{Handbook of Social Choice and Welfare}}.
  Vol.~\bibinfo{volume}{2}. \bibinfo{publisher}{Elsevier},
  \bibinfo{pages}{393--506}.
\newblock


\bibitem[\protect\citeauthoryear{Tideman}{Tideman}{1987}]%
        {tideman1987independence}
\bibfield{author}{\bibinfo{person}{T~Nicolaus Tideman}.}
  \bibinfo{year}{1987}\natexlab{}.
\newblock \showarticletitle{Independence of clones as a criterion for voting
  rules}.
\newblock \bibinfo{journal}{\emph{Social Choice and Welfare}}
  \bibinfo{volume}{4} (\bibinfo{year}{1987}), \bibinfo{pages}{185--206}.
\newblock


\bibitem[\protect\citeauthoryear{Wang, Zhang, and Yuan}{Wang
  et~al\mbox{.}}{2017}]%
        {wang2017display}
\bibfield{author}{\bibinfo{person}{Jun Wang}, \bibinfo{person}{Weinan Zhang},
  {and} \bibinfo{person}{Shuai Yuan}.} \bibinfo{year}{2017}\natexlab{}.
\newblock \bibinfo{title}{Display Advertising with Real-Time Bidding (RTB) and
  Behavioural Targeting}.
\newblock
\newblock
\showeprint[arxiv]{1610.03013}~[cs.GT]


\bibitem[\protect\citeauthoryear{Wu, Ma, Mitra, Diaz, and Liu}{Wu
  et~al\mbox{.}}{2022}]%
        {wu2022multi}
\bibfield{author}{\bibinfo{person}{Haolun Wu}, \bibinfo{person}{Chen Ma},
  \bibinfo{person}{Bhaskar Mitra}, \bibinfo{person}{Fernando Diaz}, {and}
  \bibinfo{person}{Xue Liu}.} \bibinfo{year}{2022}\natexlab{}.
\newblock \showarticletitle{A multi-objective optimization framework for
  multi-stakeholder fairness-aware recommendation}.
\newblock \bibinfo{journal}{\emph{ACM Transactions on Information Systems}}
  \bibinfo{volume}{41}, \bibinfo{number}{2} (\bibinfo{year}{2022}),
  \bibinfo{pages}{1--29}.
\newblock


\bibitem[\protect\citeauthoryear{Wu, Cao, and Xu}{Wu et~al\mbox{.}}{2023}]%
        {wu2023faster}
\bibfield{author}{\bibinfo{person}{Yao Wu}, \bibinfo{person}{Jian Cao}, {and}
  \bibinfo{person}{Guandong Xu}.} \bibinfo{year}{2023}\natexlab{}.
\newblock \showarticletitle{FASTER: A Dynamic Fairness-assurance Strategy for
  Session-based Recommender Systems}.
\newblock \bibinfo{journal}{\emph{ACM Transactions on Information Systems}}
  (\bibinfo{year}{2023}).
\newblock


\bibitem[\protect\citeauthoryear{Yuan, Abidin, Sloan, and Wang}{Yuan
  et~al\mbox{.}}{2012}]%
        {yuan2012internet}
\bibfield{author}{\bibinfo{person}{Shuai Yuan}, \bibinfo{person}{Ahmad~Zainal
  Abidin}, \bibinfo{person}{Marc Sloan}, {and} \bibinfo{person}{Jun Wang}.}
  \bibinfo{year}{2012}\natexlab{}.
\newblock \bibinfo{title}{Internet Advertising: An Interplay among Advertisers,
  Online Publishers, Ad Exchanges and Web Users}.
\newblock
\newblock
\showeprint[arxiv]{1206.1754}~[cs.IR]


\bibitem[\protect\citeauthoryear{Yuan, Wang, and Zhao}{Yuan
  et~al\mbox{.}}{2013}]%
        {yuan2013real}
\bibfield{author}{\bibinfo{person}{Shuai Yuan}, \bibinfo{person}{Jun Wang},
  {and} \bibinfo{person}{Xiaoxue Zhao}.} \bibinfo{year}{2013}\natexlab{}.
\newblock \showarticletitle{Real-time bidding for online advertising:
  measurement and analysis}. In \bibinfo{booktitle}{\emph{Proceedings of the
  Seventh International Workshop on Data Mining for Online Advertising}}. ACM,
  \bibinfo{pages}{3}.
\newblock


\bibitem[\protect\citeauthoryear{Zehlike, S{\"u}hr, Baeza-Yates, Bonchi,
  Castillo, and Hajian}{Zehlike et~al\mbox{.}}{2022}]%
        {zehlike2022fair}
\bibfield{author}{\bibinfo{person}{Meike Zehlike}, \bibinfo{person}{Tom
  S{\"u}hr}, \bibinfo{person}{Ricardo Baeza-Yates}, \bibinfo{person}{Francesco
  Bonchi}, \bibinfo{person}{Carlos Castillo}, {and} \bibinfo{person}{Sara
  Hajian}.} \bibinfo{year}{2022}\natexlab{}.
\newblock \showarticletitle{Fair Top-k Ranking with multiple protected groups}.
\newblock \bibinfo{journal}{\emph{Information Processing \& Management}}
  \bibinfo{volume}{59}, \bibinfo{number}{1} (\bibinfo{year}{2022}),
  \bibinfo{pages}{102707}.
\newblock


\bibitem[\protect\citeauthoryear{Zhang and Wang}{Zhang and Wang}{2021}]%
        {zhang2021recommendation}
\bibfield{author}{\bibinfo{person}{Dell Zhang} {and} \bibinfo{person}{Jun
  Wang}.} \bibinfo{year}{2021}\natexlab{}.
\newblock \showarticletitle{Recommendation fairness: From static to dynamic}.
\newblock \bibinfo{journal}{\emph{arXiv preprint arXiv:2109.03150}}
  (\bibinfo{year}{2021}).
\newblock


\bibitem[\protect\citeauthoryear{Zhang, Yuan, and Wang}{Zhang
  et~al\mbox{.}}{2014}]%
        {optimalbiding}
\bibfield{author}{\bibinfo{person}{Weinan Zhang}, \bibinfo{person}{Shuai Yuan},
  {and} \bibinfo{person}{Jun Wang}.} \bibinfo{year}{2014}\natexlab{}.
\newblock \showarticletitle{Optimal real-time bidding for display advertising}.
  In \bibinfo{booktitle}{\emph{Proceedings of the 20th ACM SIGKDD international
  conference on Knowledge discovery and data mining}}. ACM,
  \bibinfo{pages}{1077--1086}.
\newblock


\bibitem[\protect\citeauthoryear{Zhu, Hu, and Caverlee}{Zhu
  et~al\mbox{.}}{2018}]%
        {zhu2018fairness}
\bibfield{author}{\bibinfo{person}{Ziwei Zhu}, \bibinfo{person}{Xia Hu}, {and}
  \bibinfo{person}{James Caverlee}.} \bibinfo{year}{2018}\natexlab{}.
\newblock \showarticletitle{Fairness-aware tensor-based recommendation}. In
  \bibinfo{booktitle}{\emph{Proceedings of the 27th ACM International
  Conference on Information and Knowledge Management}}.
  \bibinfo{pages}{1153--1162}.
\newblock


\bibitem[\protect\citeauthoryear{Zwicker}{Zwicker}{2016}]%
        {Zwicker:Voting}
\bibfield{author}{\bibinfo{person}{William~S. Zwicker}.}
  \bibinfo{year}{2016}\natexlab{}.
\newblock \showarticletitle{Introduction to the Theory of Voting}.
\newblock In \bibinfo{booktitle}{\emph{Handbook of Computational Social
  Choice}}, \bibfield{editor}{\bibinfo{person}{Felix Brandt},
  \bibinfo{person}{Vincent Conitzer}, \bibinfo{person}{Ulle Endriss},
  \bibinfo{person}{J{\'{e}}r{\^{o}}me Lang}, {and} \bibinfo{person}{Ariel~D.
  Procaccia}} (Eds.). \bibinfo{publisher}{Cambridge University Press},
  \bibinfo{pages}{23--56}.
\newblock
\urldef\tempurl%
\url{https://doi.org/10.1017/CBO9781107446984.003}
\showDOI{\tempurl}


\end{thebibliography}
\end{document}